\documentclass{article}

\usepackage{amsmath,amsfonts,amssymb,amsthm}
\usepackage{microtype}
\usepackage{graphicx}
\usepackage{subfigure}
\usepackage{booktabs} 
\usepackage{xr}
\usepackage{dsfont}
\usepackage{multirow}
\usepackage{multicol}
\usepackage{color, colortbl}
\usepackage{adjustbox}
\usepackage{xspace}
\usepackage{float}

\definecolor{Gray}{gray}{0.9}

\newcommand{\bs}{YOLO-joint}
\newcommand{\bbsshort}{YOLO-ft}
\newcommand{\bbs}{YOLO-ft-full}

\usepackage{hyperref}

\newcommand{\model}{TFA\xspace}

\renewcommand{\vec}[1]{\mathbf{#1}}
\newcommand\minisection[1]{\vspace{1mm}\noindent \textbf{#1}}

\usepackage[preprint]{icml2020}

\icmltitlerunning{Frustratingly Simple Few-Shot Object Detection}

\begin{document}

\twocolumn[
\icmltitle{Frustratingly Simple Few-Shot Object Detection} 



\icmlsetsymbol{equal}{*}

\begin{icmlauthorlist}
\icmlauthor{Xin Wang}{equal,to}
\icmlauthor{Thomas E. Huang}{equal,goo}
\icmlauthor{Trevor Darrell}{to}
\icmlauthor{Joseph E. Gonzalez}{to}
\icmlauthor{Fisher Yu}{to}
\end{icmlauthorlist}

\icmlaffiliation{to}{EECS, UC Berkeley}
\icmlaffiliation{goo}{EECS, University of Michigan}

\icmlcorrespondingauthor{Xin Wang}{xinw@berkeley.edu}
\icmlcorrespondingauthor{Thomas E. Huang}{thomaseh@umich.edu}

\icmlkeywords{Machine Learning, ICML}
\vskip 0.3in
]



\begin{NoHyper}\printAffiliationsAndNotice{\icmlEqualContribution}\end{NoHyper}

\begin{abstract}
Detecting rare objects from a few examples is an emerging problem. Prior works show meta-learning is a promising approach. But, fine-tuning techniques have drawn scant attention.
We find that fine-tuning only the last layer of existing detectors on rare classes is crucial to the few-shot object detection task. Such a simple approach outperforms the meta-learning methods by roughly  2$\sim$20 points on current benchmarks and sometimes even doubles the accuracy of the prior methods. However, the high variance in the few samples often leads to the unreliability of existing benchmarks. We revise the evaluation protocols by sampling multiple groups of training examples to obtain stable comparisons and build new benchmarks based on three datasets: PASCAL VOC, COCO and LVIS. Again, our fine-tuning approach establishes a new state of the art on the revised benchmarks. The code as well as the pretrained models are available at~\url{https://github.com/ucbdrive/few-shot-object-detection}.
\end{abstract}

\section{Introduction}
Machine perception systems have witnessed significant progress in the past years.
Yet, our ability to train models that generalize to novel concepts without abundant labeled data is still far from satisfactory  when compared to human visual systems. 
Even a toddler can easily recognize a new concept with very little instruction~\cite{landau1988importance,samuelson2005they,smith2002object}.

The ability to generalize from only a few examples (so called few-shot learning) has become a key area of interest in the machine learning community.
Many \cite{vinyals2016matching,snell2017prototypical,finn2017model,hariharan2017low,gidaris2018dynamic,wang2019tafe} have explored techniques to transfer knowledge from the data-abundant base classes to the data-scarce novel classes through \emph{meta-learning}. 
They use simulated few-shot tasks by sampling from base classes during training to learn to learn from the few examples in the novel classes.

However, much of this work has focused on basic image classification tasks. 
In contrast, few-shot object detection has received far less attention. 
Unlike image classification, object detection requires the model to not only recognize the object types but also localize the targets among millions of potential regions. This additional subtask substantially
raises the overall complexity. Several~\cite{kang2019few,yan2019meta,wang2019meta} have attempted to tackle the under-explored few-shot object detection task, where only a few labeled bounding boxes are available for novel classes. These methods attach meta learners to existing object detection networks, following the meta-learning methods for classification. But, current evaluation protocols suffer from statistical unreliability, and the accuracy of baseline methods, especially simple fine-tuning, on few-object detection are not consistent in the literature.


\begin{figure*}[ht]
    \centering
    \includegraphics[width=\linewidth]{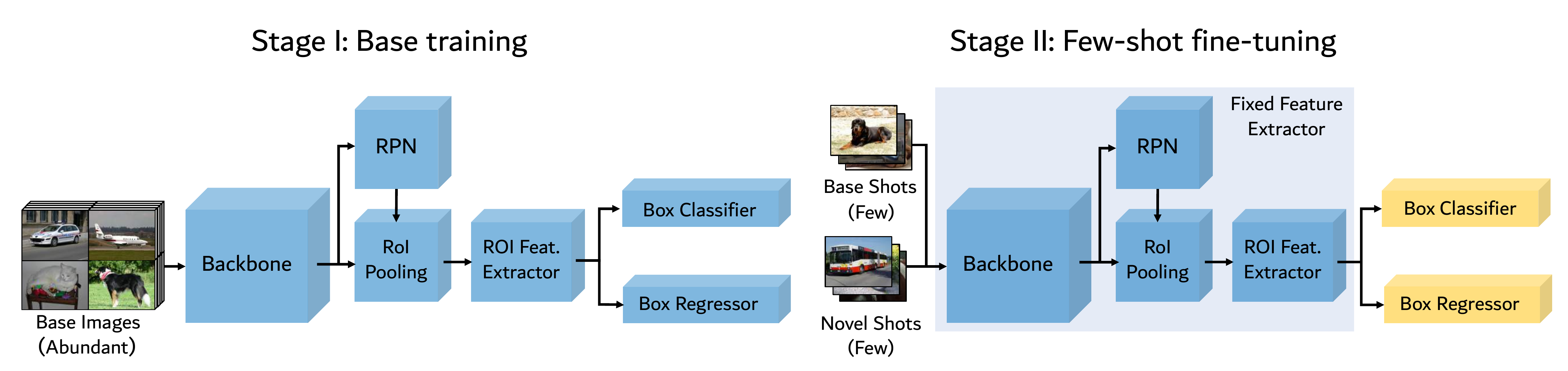}
    \vspace{-1cm}
    \caption{Illustration of our two-stage fine-tuning approach (\model). In the base training stage, the entire object detector, including both the feature extractor $\mathcal{F}$ and the box predictor, are jointly trained on the base classes. In the few-shot fine-tuning stage, the feature extractor components are fixed and only the box predictor is fine-tuned on a balanced subset consisting of both the base and novel classes.}
    \label{fig:tfa_arch}
\end{figure*}

In this work, we propose improved methods to evaluate few-shot object detection. We carefully examine fine-tuning based approaches, which are 
considered to be under-performing in the previous works~\cite{kang2019few,yan2019meta,wang2019meta}.
We focus on the training schedule and the instance-level feature normalization of the object detectors in model design and training based on fine-tuning.

We adopt a two-stage training scheme for fine-tuning as shown in Figure~\ref{fig:tfa_arch}. We first train the entire object detector, such as Faster R-CNN~\cite{ren2015faster}, on the data-abundant base classes, and then only fine-tune the last layers of the detector
on a small balanced training set consisting of both base and novel classes while freezing the other parameters of the model. 
During the fine-tuning stage, we introduce instance-level feature normalization to the box classifier inspired by~\citet{gidaris2018dynamic,qi2018low,chen2019closer}.

We find that this two-stage fine-tuning approach (\model) outperforms all previous
state-of-the-art meta-learning based methods by 2$\sim$20 points on the existing PASCAL
VOC~\cite{pascal-voc-2007} and COCO~\cite{Lin2014MicrosoftCC} benchmarks. 
When training on a single novel example (one-shot learning), our method can achieve twice
the accuracy of prior sophisticated state-of-the-art approaches.

Several issues with the existing evaluation protocols prevent consistent model comparisons. The accuracy measurements have high variance, making published comparisons unreliable.  Also, the previous evaluations only report the detection accuracy on the novel classes, and fail to evaluate knowledge retention on the base classes.

To resolve these issues, we build new benchmarks on three datasets: PASCAL VOC, COCO and LVIS~\cite{gupta2019lvis}.
We sample different groups of few-shot training examples for multiple runs of the experiments to obtain a stable accuracy estimation and quantitatively analyze the variances of different evaluation metrics. The new evaluation reports the average precision (AP) on both the base classes and novel classes as well as the mean AP on all classes, referred to as the generalized few-shot learning setting in the few-shot classification literature~\cite{hariharan2017low,wang2019tafe}.

Our fine-tuning approach establishes new states of the art on the benchmarks.  
On the challenging LVIS dataset, our two-stage training scheme improves the average detection precision of rare classes
($<$10 images) by $\sim$4 points and common classes (10-100 images) by $\sim$2 points with negligible precision loss for the frequent classes ($>$100 images). 

\section{Related Work}
Our work is related to the rich literature on few-shot image classification, which uses various 
meta-learning based or metric-learning based methods. We also draw connections between our work and the existing meta-learning based few-shot object detection methods. To the best of our knowledge, we are the first to conduct a systematic analysis of fine-tuning based approaches on few-shot object detection. 

\minisection{Meta-learning.} The goal of meta-learning is to acquire task-level meta knowledge that
can help the model quickly adapt to new tasks and environments with very few labeled examples.  Some~\cite{finn2017model,rusu2018meta,nichol2018reptile} learn to fine-tune and aim to obtain a good parameter initialization that can adapt to
new tasks with a few scholastic gradient updates. Another popular line of research on meta-learning is to use parameter generation during adaptation to novel tasks. \citet{gidaris2018dynamic} propose an attention-based weight generator to generate the classifier weights for the novel classes. \citet{wang2019tafe} construct task-aware feature embeddings by generating parameters for the feature layers. These approaches have only been used for few-shot image 
classification and not on more challenging tasks like object detection.

However, some~\cite{chen2019closer} 
raise concerns about the reliability of the results given 
that a consistent comparison of different approaches is missing. 
Some simple fine-tuning based approaches, which draw little attention in the
community, turn out to be more favorable than many prior works that use meta-learning
on few-shot image classification~\cite{chen2019closer,dhillon2019baseline}.
As for the emerging few-shot object detection task, there is neither consensus on the evaluation benchmarks nor a consistent comparison of different approaches due to the increased network complexity, obscure implementation details, and variances in evaluation protocols.

\minisection{Metric-learning.} Another line of work~\cite{koch2015siamese,snell2017prototypical,vinyals2016matching}
focuses on learning to compare or metric-learning. Intuitively, if the model can construct distance metrics to 
estimate the similarity between two input images, it may generalize to 
novel categories with few labeled instances. More recently, several~\cite{chen2019closer,gidaris2018dynamic,qi2018low} adopt a 
cosine similarity based classifier to reduce the intra-class variance on the few-shot classification task, which leads to favorable performance compared to many 
meta-learning based approaches. Our method also adopts a cosine
similarity classifier to classify the categories of the region proposals. However, we focus on the instance-level distance measurement rather than on the image level.

\minisection{Few-shot object detection.} There are several early attempts 
at few-shot object detection using meta-learning. ~\citet{kang2019few} and 
~\citet{yan2019meta} apply feature re-weighting schemes to a
single-stage object detector (YOLOv2) and a two-stage object detector
(Faster R-CNN), with the help of a meta learner that takes the support images (\textit{i.e.}, a small number of labeled images of the novel/base classes) 
as well as the bounding box annotations as inputs. ~\citet{wang2019meta} 
propose a weight prediction meta-model to predict parameters of
category-specific components from the few examples while learning the category-agnostic components from base class examples. 

In all these works, fine-tuning based approaches are considered as baselines
with worse performance than meta-learning based approaches. They consider
\emph{jointly fine-tuning}, where base classes and novel classes are trained together, and \emph{fine-tuning the entire model}, where the detector is first trained on the base classes only and then fine-tuned on a balanced set with both base and novel classes. In contrast, we find that fine-tuning only the last layer of the object detector on the balanced subset and keeping the rest of model fixed can substantially improve the
detection accuracy, outperforming all the prior meta-learning based
approaches. This indicates that feature representations
learned from the base classes might be able to transfer to the novel classes
and simple adjustments to the box predictor can provide 
strong performance gain~\cite{dhillon2019baseline}.

\vspace{-0.3cm}
\section{Algorithms for Few-Shot Object Detection}

In this section, we start with the preliminaries on the few-shot object detection setting. Then, we talk about our two-stage fine-tuning approach in
Section~\ref{sec:tfa}. Section~\ref{sec:meta} summarizes the previous meta-learning approaches.

We follow the few-shot object detection settings introduced in~\citet{kang2019few}. There are a set of base
classes $C_b$ that have many instances and a set of novel classes $C_n$ that
have only $K$ (usually less than 10) instances per category.
For an object detection dataset $\mathcal{D}=\{(x, y), x\in\mathcal{X}, y\in\mathcal{Y}\}$, where $x$
is the input image and $y=\{(c_i, \vec{l}_i), i=1, ...,N\}$ denotes the categories $c \in C_b \cup C_n$
and bounding box coordinates $\vec{l}$ of the $N$ object instances in the image $x$.
For synthetic few-shot datasets using
PASCAL VOC and COCO, the novel set for training is balanced and each class has the same number of 
annotated objects (\textit{i.e.}, $K$-shot). The recent LVIS dataset has a natural long-tail distribution, which does
not have the manual $K$-shot split. The classes in LVIS are divided into \emph{frequent} classes
(appearing in more than 100 images), \emph{common} classes (10-100 images), and \emph{rare}
classes (less than 10 images). We consider both synthetic and natural datasets in our work and
follow the naming convention of $k$-shot for simplicity. 

The few-shot object detector is evaluated on a test set of both the base classes and
the novel classes. The goal is to optimize the detection accuracy measured by average precision (AP)
of the novel classes as well as the base classes. This setting is different from the $N$-way-$K$-shot setting~\cite{finn2017model,vinyals2016matching,snell2017prototypical}
commonly used in few-shot classification. 

\subsection{Two-stage fine-tuning approach}
\label{sec:tfa}
We describe our two-stage fine-tuning approach (\model) for few-shot object detection in
this section. We adopt the widely used Faster R-CNN~\cite{ren2015faster}, a two-stage object
detector, as our base detection model. As shown in Figure~\ref{fig:tfa_arch}, the feature learning components, referred to as $\mathcal{F}$, of a Faster R-CNN model include the backbone (\textit{e.g.}, ResNet~\cite{he2016deep}, VGG16~\cite{simonyan2014very}), the region proposal network (RPN), as well as a two-layer
fully-connected (FC) sub-network as a proposal-level feature extractor. 
There is also a box predictor composed of a box classifier $\mathcal{C}$ to classify the
object categories and a box regressor $\mathcal{R}$ to predict the bounding box coordinates. 
Intuitively, the backbone features as well as the RPN features
are class-agnostic. Therefore, features learned from the base classes are likely to transfer
to the novel classes without further parameter updates. The key component of our method is to separate the
feature representation learning and the box predictor learning into two stages. 

\minisection{Base model training.} In the first stage, we train the feature extractor 
and the box predictor only on the base classes $C_b$, with the same loss function used in ~\citet{ren2015faster}. The joint loss is, 
\begin{equation}
    \mathcal{L} = \mathcal{L}_{\text{rpn}} + \mathcal{L}_{\text{cls}} + \mathcal{L}_{\text{loc}},
    \label{eq:loss}
\end{equation}
where $\mathcal{L}_\text{rpn}$ is applied
to the output of the RPN to distinguish foreground from backgrounds and refine the
anchors, $\mathcal{L}_\text{cls}$ is a cross-entropy loss for the box classifier $\mathcal{C}$,
and $\mathcal{L}_\text{loc}$ is a smoothed $L_1$ loss for the box regressor $\mathcal{R}$.

\minisection{Few-shot fine-tuning.} In the second stage, we create a small balanced training set 
with $K$ shots per class, containing both base and novel classes.
We assign randomly initialized weights to the box prediction networks for the novel classes
and fine-tune only the box classification and
regression networks, namely the last layers of the detection model, while keeping the entire 
feature extractor $\mathcal{F}$ fixed. We use the same loss function
in Equation~\ref{eq:loss} and a smaller learning rate. The learning rate is reduced by 20 from the first stage in all our experiments. 

\begin{figure*}[ht]
    \centering
    \includegraphics[width=\linewidth]{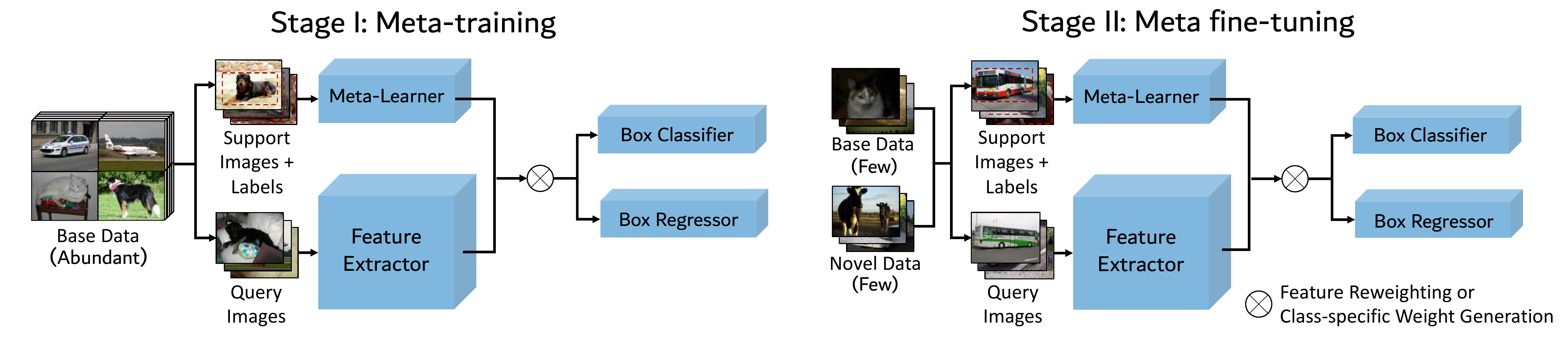}
    \vspace{-8mm}
    \caption{Abstraction of the meta-learning based few-shot object detectors. A meta-learner is introduced to acquire task-level meta information and help the model generalize to novel classes through feature re-weighting (\textit{e.g.}, FSRW and Meta R-CNN) or weight generation (\textit{e.g.}, MetaDet). A two-stage training approach (meta-training and meta fine-tuning) with episodic learning is commonly adopted.}
    \label{fig:meta_arch}
\end{figure*}

\minisection{Cosine similarity for box classifier.} We consider using a classifier based on cosine similarity in the second fine-tuning stage, inspired by ~\citet{gidaris2018dynamic,qi2018low,chen2019closer}. The weight matrix $W\in\mathbb{R}^{d\times c}$ of the box classifier $\mathcal{C}$ can be written as $[w_1, w_2, ..., w_c]$, where $w_c\in\mathbb{R}^d$ is the per-class weight vector. The output of $\mathcal{C}$ is scaled similarity scores $S$ of the input feature $\mathcal{F}(x)$ and the weight vectors of different classes. The entries in $S$ are 
\begin{equation}
    s_{i,j} = \frac{\alpha \mathcal{F}(x)_i^\top w_j}{\|\mathcal{F}(x)_i\| \|w_j\|}, 
\end{equation}
where $s_{i,j}$ is the similarity score between the $i$-th object proposal of the input $x$ and the weight vector of class $j$. $\alpha$ is the scaling factor. We use a fixed $\alpha$ of 20 in our experiments. We find empirically that the instance-level feature 
normalization used in the cosine similarity based classifier helps reduce the intra-class
variance and improves the detection accuracy of novel classes with less decrease in detection accuracy of base classes when compared to a FC-based classifier, especially when the number of training examples is small.

\subsection{Meta-learning based approaches}
\label{sec:meta}
We describe the existing meta-learning based few-shot object detection networks, including FSRW~\cite{kang2019few}, Meta R-CNN~\cite{yan2019meta} and MetaDet~\cite{wang2019meta}, in 
this section to draw comparisons with our approach.
Figure~\ref{fig:meta_arch} illustrates the structures of these networks.
In meta-learning approaches, in addition to the base object detection model that is either single-stage or 
two-stage, a meta-learner is introduced to acquire class-level meta knowledge and
help the model generalize to novel classes through feature re-weighting, such as FSRW
and Meta R-CNN, or class-specific weight generation, such as MetaDet. The input to the
meta learner is a small set of support images with the bounding box annotations of the target objects. 

The base object detector and the meta-learner are often jointly trained using episodic training~\cite{vinyals2016matching}.
Each episode is composed of a supporting set of $N$ objects and a set of query images.
In FSRW and Meta R-CNN, the support images and the binary masks of the annotated objects are used as input to the meta-learner, which generates class reweighting vectors that modulate the feature representation of the query images.
As shown in Figure~\ref{fig:meta_arch}, the training procedure is also split into a meta-training stage, where the model is only trained on the data of the base classes, and a meta fine-tuning stage, where the support set includes the few examples of the novel classes and a subset of examples from the base classes.

Both the meta-learning approaches and our approach have a two-stage 
training scheme. However, we find that the episodic learning used in meta-learning approaches 
can be very memory inefficient as the number of classes in the supporting set increases. Our
fine-tuning method only fine-tunes the last layers of the network with a normal batch training scheme, which is much more memory efficient. 

\section{Experiments}
\begin{table*}[ht]
\centering
\footnotesize
\setlength{\tabcolsep}{0.4em}
\caption{{Few-shot detection performance (mAP50) on the PASCAL VOC dataset.} We evaluate the performance on three different sets of novel classes. Our approach consistently outperforms baseline methods by a large margin (about 2$\sim$20 points), especially when the number of shots is low. FRCN stands for Faster R-CNN. \model w/ cos is our approach with a cosine similarity based box classifier. \vspace{1mm}
}
\adjustbox{width=\linewidth}{
\begin{tabular}{l|c|ccccc|ccccc|ccccc}
\toprule
\multirow{2}{*}{Method / Shot} & \multirow{2}{*}{Backbone} & \multicolumn{5}{c|}{Novel Set 1} & \multicolumn{5}{c|}{Novel Set 2} & \multicolumn{5}{c}{Novel Set 3} \\ 
&  & 1     & 2     & 3    & 5    & 10   & 1     & 2     & 3    & 5    & 10   & 1     & 2     & 3    & 5    & 10   \\ \midrule
\bs{}~\cite{kang2019few}   &  \multirow{5}{*}{YOLOv2}  & 0.0   & 0.0   & 1.8  & 1.8  & 1.8  & 0.0   & 0.1   & 0.0  & 1.8  & 0.0    & 1.8   & 1.8   & 1.8  & 3.6  & 3.9  \\ 
\bbsshort{}~\cite{kang2019few}  & & 3.2   & 6.5   & 6.4  & 7.5  & 12.3 & 8.2   & 3.8   & 3.5  & 3.5  & 7.8  & 8.1   & 7.4   & 7.6  & 9.5  & 10.5 \\ 
\bbs{}~\cite{kang2019few}  &  & 6.6   & 10.7  & 12.5 & 24.8 & 38.6 & 12.5  & 4.2   & 11.6 & 16.1 & 33.9 & 13.0  & 15.9  & 15.0 & 32.2 & 38.4 \\ 
FSRW~\cite{kang2019few}  &  & 14.8  & 15.5  & 26.7 & 33.9 & 47.2 & 15.7  & 15.3  & 22.7 & 30.1 & 40.5 & 21.3  & 25.6  & 28.4 & 42.8 & 45.9 \\ 
MetaDet~\cite{wang2019meta} & & 17.1 & 19.1 & 28.9 & 35.0 & 48.8 & 18.2 & 20.6 & 25.9 & 30.6 & 41.5 & 20.1 & 22.3 & 27.9 & 41.9 & 42.9 \\ \midrule
FRCN+joint~\cite{wang2019meta} & \multirow{3}{*}{FRCN w/VGG16} & 0.3 & 0.0 & 1.2 & 0.9 & 1.7 & 0.0 & 0.0 & 1.1 & 1.9 & 1.7 & 0.2 & 0.5 & 1.2 & 1.9 & 2.8\\
FRCN+joint-ft~\cite{wang2019meta} & & 9.1 & 10.9 & 
13.7 & 25.0 & 39.5 & 10.9 & 13.2 & 17.6 & 19.5 & 36.5 & 15.0 & 15.1 & 18.3 & 33.1 & 35.9 \\
MetaDet~\cite{wang2019meta} &  & 18.9 & 20.6 & 30.2 & 36.8 & 49.6 & 21.8 & 23.1 & 27.8 & 31.7 & 43.0 & 20.6 & 23.9 & 29.4 & 43.9 & 44.1 \\ \midrule
FRCN+joint~\cite{yan2019meta} & \multirow{4}{*}{FRCN w/R-101} & 2.7 & 3.1 & 4.3 & 11.8 & 29.0 & 1.9 & 2.6 & 8.1 & 9.9 & 12.6 & 5.2 & 7.5 & 6.4 & 6.4 & 6.4 \\
FRCN+ft~\cite{yan2019meta} &  & 11.9 & 16.4 & 
29.0 & 36.9 & 36.9 & 5.9 & 8.5 & 23.4 & 29.1 & 28.8 & 5.0 & 9.6 & 18.1 & 30.8 & 43.4 \\
FRCN+ft-full~\cite{yan2019meta} &  & 13.8 & 19.6 
& 32.8 & 41.5 & 45.6 & 7.9 & 15.3 & 26.2 & 31.6 & 39.1 & 9.8 & 11.3 & 19.1 & 35.0 & 45.1 \\
Meta R-CNN~\cite{yan2019meta} &  & 19.9 & 25.5 & 35.0 & 45.7 & 51.5 & 10.4 & 19.4 & 29.6 & 34.8 & \textbf{45.4} & 14.3 & 18.2 & 27.5 & 41.2 & 48.1 \\ \midrule
FRCN+ft-full (Our Impl.) &  & 15.2 & 20.3 & 29.0 & 40.1 & 45.5 & 13.4 & 20.6 & 28.6 & 32.4 & 38.8 & 19.6 & 20.8 & 28.7 & 42.2 & 42.1 \\
\rowcolor{Gray} \model w/ fc (Ours) &  & 36.8 & 29.1 & 43.6 & \textbf{55.7} & \textbf{57.0} & 18.2 & \textbf{29.0} & 33.4 & \textbf{35.5} & 39.0 & 27.7 & 33.6 & 42.5 & 48.7 & \textbf{50.2} \\
\rowcolor{Gray} \model w/ cos (Ours) & \multirow{-3}{*}{FRCN w/R-101} & \textbf{39.8} & \textbf{36.1} & \textbf{44.7} & \textbf{55.7} & 56.0 & \textbf{23.5} & 26.9 & \textbf{34.1} & 35.1 & 39.1 & \textbf{30.8} & \textbf{34.8} & \textbf{42.8} & \textbf{49.5} & 49.8 \\
\bottomrule
\end{tabular}}
\label{tab:main_voc}
\end{table*}
In this section, we conduct extensive comparisons with previous methods on the existing 
few-shot object detection benchmarks using PASCAL VOC and COCO, where our approach can obtain
about 2$\sim$20 points improvement in all settings (Section~\ref{sec:exist_benchmark}). We then introduce a new benchmark on three datasets (PASCAL VOC, COCO and LVIS) with revised evaluation protocols to address the unreliability of previous benchmarks (Section~\ref{sec:revised_bench}). We also provide various ablation studies and visualizations in Section~\ref{sec:vis}. 

\minisection{Implementation details.}
We use Faster R-CNN~\cite{ren2015faster} as our base detector and Resnet-101~\cite{he2016deep} with a Feature Pyramid Network~\cite{lin2016feature} as the backbone.
All models are trained using SGD with a mini-batch size of 16, momentum of 0.9, and weight decay of 0.0001.
A learning rate of 0.02 is used during base training and 0.001 during few-shot fine-tuning.
For more details, the code is available at~\url{https://github.com/ucbdrive/few-shot-object-detection}.

\subsection{Existing few-shot object detection benchmark}
\label{sec:exist_benchmark}
\minisection{Existing benchmarks.} 
Following the previous work~\cite{kang2019few,yan2019meta,wang2019meta}, we 
first evaluate our approach on PASCAL VOC 2007+2012 and COCO, using the same data splits and training examples provided by ~\citet{kang2019few}.
For the few-shot PASCAL VOC dataset,  the 20 classes are randomly divided into 15 base classes and 5 novel classes, where the novel classes have $K=1,2,3,5,10$ objects per class sampled from the combination of the trainval sets of the 2007 and 2012 versions for training. Three random split groups are considered in this work. PASCAL VOC 2007 test set is used for evaluation. For COCO, the 60 categories disjoint with PASCAL VOC are used as base classes while the remaining 20 classes are used as novel classes with $K=10, 30$.  For evaluation metrics, AP50 (matching threshold is 0.5) of the novel classes is used on PASCAL VOC and the COCO-style AP of the novel classes is used on COCO. 

\minisection{Baselines.} We compare our approach with the meta-learning approaches \texttt{FSRW}, \texttt{Meta-RCNN} and \texttt{MetaDet} together with the fine-tuning
based approaches:  \emph{jointly training}, denoted by \texttt{FRCN/YOLO+joint}, where the base and novel class examples are jointly trained in one stage,  and \emph{fine-tuning the entire model}, denoted by \texttt{FRCN/YOLO+ft-full}, where both the feature extractor $\mathcal{F}$ and the box predictor ($\mathcal{C}$ and $\mathcal{R}$) are jointly fine-tuned until convergence in the second fine-tuning stage. \texttt{FRCN} is Faster R-CNN for short. Fine-tuning with less iterations, denoted by \texttt{FRCN/YOLO+ft}, are reported in ~\citet{kang2019few} and ~\citet{yan2019meta}.


\begin{table}[ht]
\centering
\footnotesize
\setlength{\tabcolsep}{0.4em}
\caption{{Few-shot detection performance for the base and novel classes on Novel Set 1 of the PASCAL VOC dataset.} Our approach outperforms baselines on both base and novel classes and does not degrade the performance on the base classes greatly.\vspace{1mm}}
\adjustbox{width=\linewidth}{
\begin{tabular}{cc|cc}
\toprule
\# shots & Method  &  Base AP50 & Novel AP50  \\ \midrule
\multirow{5}{*}{3} & FRCN+ft-full~\cite{yan2019meta} & 63.6 & 32.8\\
                  & Meta R-CNN~\cite{yan2019meta} & 64.8 & 35.0 \\ 
                  & Train base only (Our Impl.) &\textbf{80.8} & 9.0 \\
                  & FRCN+ft-full (Our Impl.) & 66.1 & 29.0 \\
                     &\cellcolor{Gray} \model w/ cos (Ours)    & \cellcolor{Gray}79.1 & \cellcolor{Gray}\textbf{44.7} \\  \midrule\midrule
\multirow{5}{*}{10} & FRCN+ft-full~\cite{yan2019meta} & 61.3 & 45.6 \\
                    & Meta R-CNN~\cite{yan2019meta} & 67.9 & 51.5 \\ 
                    & Train base only & \textbf{80.8} & 9.0 \\
                    & FRCN+ft-full (Our Impl.) & 66.0 & 45.5 \\
                    & \cellcolor{Gray}\model w/ cos (Ours) & \cellcolor{Gray}78.4 & \cellcolor{Gray}\textbf{56.0} \\
\bottomrule
\end{tabular}}
\label{tab:voc_base}
\end{table}

\minisection{Results on PASCAL VOC.} We provide the average AP50 of the novel classes on PASCAL VOC with three random splits in Table~\ref{tab:main_voc}. Our approach uses ResNet-101 as the backbone, similar to Meta R-CNN. We implement \texttt{FRCN+ft-full} in our framework, which roughly 
matches the results reported in ~\citet{yan2019meta}. \texttt{MetaDet} uses VGG16 as the backbone, but the performance is similar and sometimes worse compared to Meta R-CNN.

\begin{figure*}[ht]
	\begin{center}
		\centerline{\includegraphics[width=\columnwidth*2]{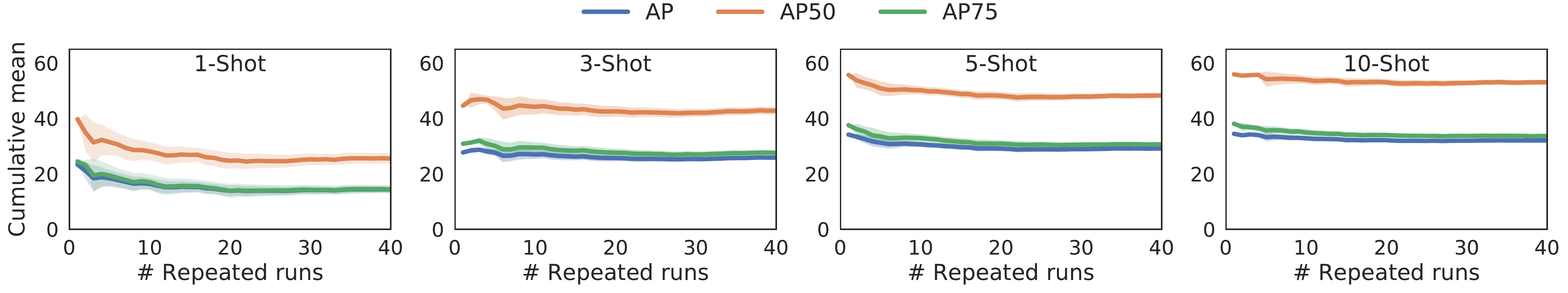}}
		\vspace{-5mm}
		\caption{Cumulative means with 95\% confidence intervals across 40 repeated runs, computed on the novel classes of the first split of PASCAL VOC. The means and variances become stable after around 30 runs.}
		\label{fig:avg-ap}
	\end{center}
	\vspace{-10mm}
\end{figure*}

In all different data splits and different numbers of training shots, our approach (the last row) is able to outperform the
previous methods by a large margin. It even doubles the performance of the previous approaches in the one-shot cases. The 
improvements, up to 20 points, is much larger than the gap among the previous meta-learning based approaches, indicating the effectiveness of our approach.  We also
compare the cosine similarity based box classifier (\texttt{\model+w/cos}) with a normal FC-based classifier (\texttt{\model+w/fc}) and find that \texttt{\model+w/cos} is better than \texttt{\model+w/fc} on extremely low shots (\textit{e.g.,} 1-shot), but the two are roughly similar when there are more training shots, \textit{e.g.}, 10-shot. 

For more detailed comparisons, we cite the numbers from~\citet{yan2019meta} of their model performance on the base classes in Table~\ref{tab:voc_base}. We find that our model has a much higher average AP on the base classes than \texttt{Meta R-CNN} with a gap of about 10 to 15 points. To eliminate the differences in implementation details, we also report our re-implementation of \texttt{FRCN+ft-full} and \texttt{training base only}, where the base classes should have the highest accuracy as the model is only trained on the base classes examples. We find that our model has a small decrease in performance, less than 2 points, on the base classes.


\begin{table}[ht]
    \centering
    \footnotesize
\caption{{Few-shot detection performance for the novel categories on the COCO dataset.} Our approach consistently outperforms baseline methods across all shots and metrics.\vspace{2mm}}
    \adjustbox{width=\linewidth}{
    \begin{tabular}{l|cc|cc}
    \toprule
    \multirow{2}{*}{Model} 
    &\multicolumn{2}{c|}{novel AP} & \multicolumn{2}{c}{novel AP75}\\
    & 10 & 30 & 10 & 30 \\ \midrule
    FSRW\small{~\cite{kang2019few}}  & 5.6 & 9.1 & 4.6 & 7.6 \\ 
    MetaDet\small{~\cite{wang2019meta}} & 7.1 & 11.3 & 6.1 & 8.1 \\
    FRCN+ft+full\small{~\cite{yan2019meta}} &  6.5 & 11.1 & 5.9 & 10.3\\
    Meta R-CNN\small{~\cite{yan2019meta}}  & 8.7 & 12.4 & 6.6 & 10.8\\ 
    FRCN+ft-full (Our Impl.)  & 9.2 & 12.5  & 9.2 & 12.0  \\ 	
    \rowcolor{Gray} \model w/ fc (Ours)  & \textbf{10.0} & 13.4  & 9.2 & 13.2 \\
    \rowcolor{Gray} \model w/ cos (Ours) & \textbf{10.0} & \textbf{13.7}  & \textbf{9.3} & \textbf{13.4}\\
    \bottomrule
    \end{tabular}}
\label{tab:main_coco}
\end{table}

\minisection{Results on COCO.} Similarly, we report the average AP and AP75 of the 20 novel classes on COCO in Table~\ref{tab:main_coco}. AP75 means matching threshold is 0.75, a more strict metric than AP50. Again, we consistently outperform previous methods across all shots on both novel AP and
novel AP75. We achieve around 1 point improvement in AP over the best performing baseline and around 2.5 points improvement in AP75.

\begin{table*}[t]
	\centering
	\footnotesize
	\setlength{\tabcolsep}{0.4em}
	\caption{Generalized object detection benchmarks on LVIS. We compare our approach to the baselines provided in LVIS~\cite{gupta2019lvis}. Our approach outperforms the corresponding baseline across all metrics, backbones, and sampling schemes. \vspace{1mm}}
	\adjustbox{width=.8\linewidth}{
		\begin{tabular}{c|c|c|ccc|ccc|ccc}
			\toprule
			Method & Backbone  & Repeated sampling & AP & AP50 & AP75 & APs & APm & APl & APr & APc & APf \\\midrule
			Joint training~\cite{gupta2019lvis}  & \multirow{3}{*}{FRCN w/ R-50} && 19.8 & 33.6 & 20.4 & 17.1 & 25.9 & 33.2 & 2.1 & 18.5 & \textbf{28.5}  \\
			\model w/ fc (Ours) &  &  & 22.3 & 37.8 & 22.2 & 18.5 & 28.2 & 36.6 & 14.3 & 21.1 & 27.0 \\
			\model w/ cos (Ours) &  &  & \cellcolor{Gray} 22.7 & \cellcolor{Gray} 37.2 & \cellcolor{Gray} 23.9 & \cellcolor{Gray} 18.8 & \cellcolor{Gray} 27.7 & \cellcolor{Gray} 37.1 & \cellcolor{Gray} 15.4 & \cellcolor{Gray} 20.5 & \cellcolor{Gray} 28.4 \\ \midrule
			Joint training~\cite{gupta2019lvis}  & \multirow{3}{*}{FRCN w/ R-50}  &\multirow{3}{*}{\checkmark} & 23.1 & 38.4 & 24.3 & 18.1 & 28.3 & 36.0 & 13.0 & 22.0 & 28.4 \\ 
			\model w/ fc (Ours) &  & & 24.1 & 39.9 & 25.4 & 19.5 & 29.1 & 36.7 & 14.9 & 23.9 & 27.9 \\
			\model w/ cos (Ours) & &&\cellcolor{Gray} \textbf{24.4} & \cellcolor{Gray}\textbf{40.0} & \cellcolor{Gray}\textbf{26.1} & \cellcolor{Gray}\textbf{19.9} & \cellcolor{Gray}\textbf{29.5} & \cellcolor{Gray}\textbf{38.2} &\cellcolor{Gray}\textbf{16.9} & \cellcolor{Gray}\textbf{24.3} & \cellcolor{Gray}27.7 \\ \midrule\midrule
			Joint training~\cite{gupta2019lvis}  & \multirow{3}{*}{FRCN w/ R-101} & & 21.9 & 35.8 & 23.0 & 18.8 & 28.0 & 36.2 & 3.0 & 20.8 & \textbf{30.8}  \\
			\model w/ fc (Ours) &  & & 23.9 & 39.3 & 25.3 & 19.5 & 29.5 & 38.6 & 16.2 & 22.3 & 28.9  \\
			\model w/ cos (Ours) &  & & \cellcolor{Gray} 24.3 & \cellcolor{Gray} 39.3 & \cellcolor{Gray} 25.8 & \cellcolor{Gray} 20.1 & \cellcolor{Gray} 30.2 & \cellcolor{Gray} 39.5 & \cellcolor{Gray} \textbf{18.1} & \cellcolor{Gray} 21.8 & \cellcolor{Gray} 29.8
  \\\midrule
			Joint training~\cite{gupta2019lvis} & \multirow{3}{*}{FRCN w/ R-101}  & \multirow{3}{*}{\checkmark} & 24.7 & 40.5 & 26.0 & 19.0 & 30.3 & 38.0 & 13.4 & 24.0 & 30.1 \\ 
			\model w/ fc (Ours) &  & & 25.4 & \textbf{41.8} & 27.0 & 19.8 & 31.1 & 39.2 & 15.5 & 26.0 & 28.6 \\
			\model w/ cos (Ours) & & & \cellcolor{Gray} \textbf{26.2} & \cellcolor{Gray}\textbf{41.8} & \cellcolor{Gray}\textbf{27.5} & \cellcolor{Gray}\textbf{20.2} & \cellcolor{Gray}\textbf{32.0} & \cellcolor{Gray}\textbf{39.9} & \cellcolor{Gray}17.3 & \cellcolor{Gray}\textbf{26.4} & \cellcolor{Gray}29.6 \\ 
			\bottomrule
	\end{tabular}}
	\label{tab:lvis_bench}
\end{table*}

\begin{figure*}[ht]
	\begin{center}
		\centerline{\includegraphics[width=\linewidth]{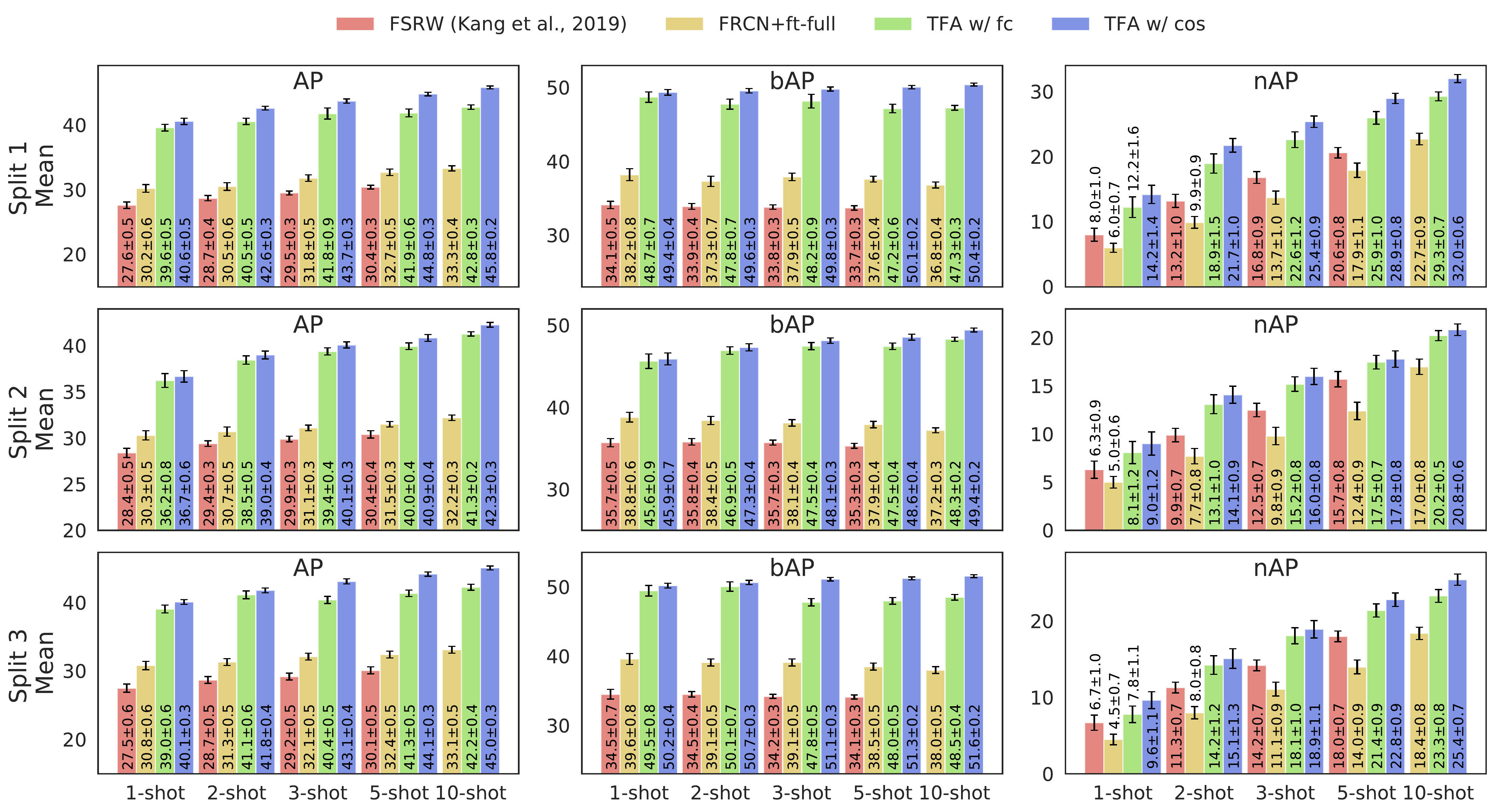}}
		\vspace{-5mm}
		\caption{Generalized object detection benchmarks on PASCAL VOC. For each metric, we report the average and 95\% confidence interval computed over 30 random samples.}
		\label{fig:voc_bench}
	\end{center}
	\vspace{-10mm}
\end{figure*}

\subsection{Generalized few-shot object detection benchmark}
\label{sec:revised_bench}
\minisection{Revised evaluation protocols.}
We find several issues with existing benchmarks.
First, previous evaluation protocols focus only on the performance on novel classes.
This ignores the potential performance drop in base classes and thus the overall performance of the network.
Second, the sample variance is large due to the few samples that are used for training.
This makes it difficult to draw conclusions from comparisons against other methods, as differences in performance could be insignificant.

To address these issues, we first revise the evaluation protocol to include evaluation on base classes. On our benchmark, we report AP on base classes (bAP) and the overall AP in addition to AP on the novel classes (nAP). This allows us to observe trends in performance on both base and novel classes, and the overall performance of the network.

Additionally, we train our models for multiple runs on different random samples of training shots to obtain averages and confidence intervals.
In Figure~\ref{fig:avg-ap}, we show the cumulative mean and 95\% confidence interval across 40 repeated runs with $K=1,3,5,10$ on the first split of PASCAL VOC.
Although the performance is high on the first random sample, the average decreases significantly as more samples are used.
Additionally, the confidence intervals across the first few runs are large, especially in the low-shot scenario.
When we use more repeated runs, the averages stabilizes and the confidence intervals become small, which allows for better comparisons.

\minisection{Results on LVIS.}
We evaluate our approach on the recently introduced LVIS dataset~\cite{gupta2019lvis}. The number of images in each category in LVIS has a natural long-tail distribution. We treat the frequent and common classes as base classes, and the rare classes as novel classes.
The base training is the same as before.
During few-shot fine-tuning, we artificially create a balanced subset of the entire dataset by sampling up to 10 instances for each class and fine-tune on this subset.

We show evaluation results on LVIS in Table~\ref{tab:lvis_bench}.
Compared to the methods in~\citet{gupta2019lvis}, our approach is able to achieve better performance of $\sim$1-1.5 points in overall AP and $\sim$2-4 points in AP for rare and common classes.
We also demonstrate results without using repeated sampling, which is a weighted sampling scheme that is used in~\citet{gupta2019lvis} to address the data imbalance issue.
In this setting, the baseline methods can only achieve $\sim$2-3 points in AP for rare classes.
On the other hand, our approach is able to greatly outperform the baseline and increase the AP on rare classes by around 13 points and on common classes by around 1 point.
Our two-stage fine-tuning scheme is able to address the severe data imbalance issue without needing repeated sampling.

\begin{figure*}[ht]
	\begin{center}
		\centerline{\includegraphics[width=.9\linewidth]{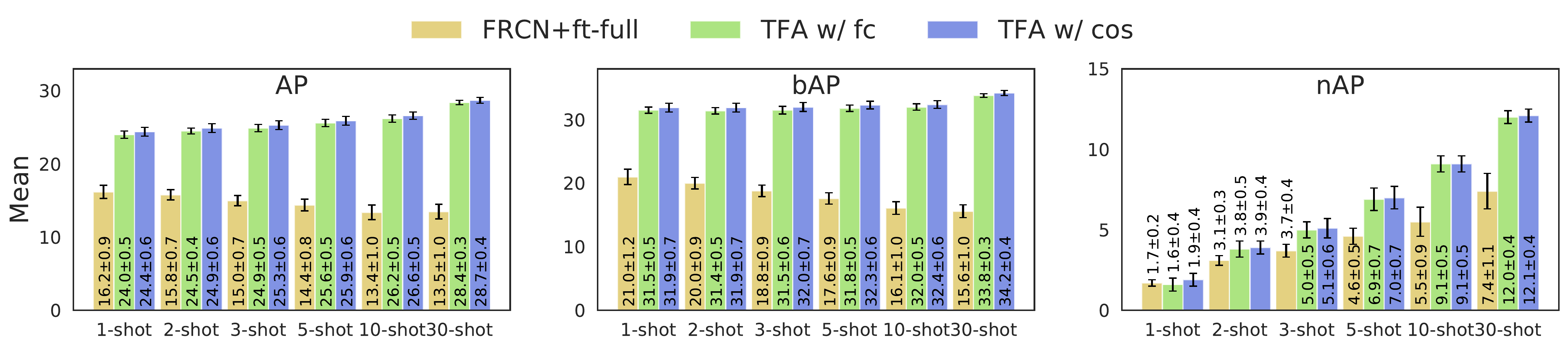}}
		\vspace{-5mm}
		\caption{Generalized object detection benchmarks on COCO. For each metric, we report the average and 95\% confidence interval computed over 10 random samples.}
		\label{fig:coco_bench}
	\end{center}
\end{figure*}

\vspace{3mm}
\minisection{Results on PASCAL VOC and COCO.}
We show evaluation results on generalized PASCAL VOC in Figure~\ref{fig:voc_bench} and COCO in Figure~\ref{fig:coco_bench}.
On both datasets, we evaluate on the base classes and the novel classes and report AP scores for each.
On PASCAL VOC, we evaluate our models over 30 repeated runs and report the average and the 95\% confidence interval.
On COCO, we provide results on 1, 2, 3, and 5 shots in addition to the 10 and 30 shots used by the existing benchmark for a better picture of performance trends in the low-shot regime. For the full quantitative results of other metrics (e.g., AP50 and AP75), more details are available in the appendix.


{
\begin{figure*}[!ht]
	\centering
	\footnotesize
	\setlength{\tabcolsep}{0.1em}
	\adjustbox{width=.95\linewidth}{
		\begin{tabular}{ccccccc}
		    \multirow{2}{*}{\rotatebox{90}{\hspace{4mm}PASCAL VOC}}& 
		    \rotatebox{90}{\hspace{4mm}Success} &
			\includegraphics[width=1in]{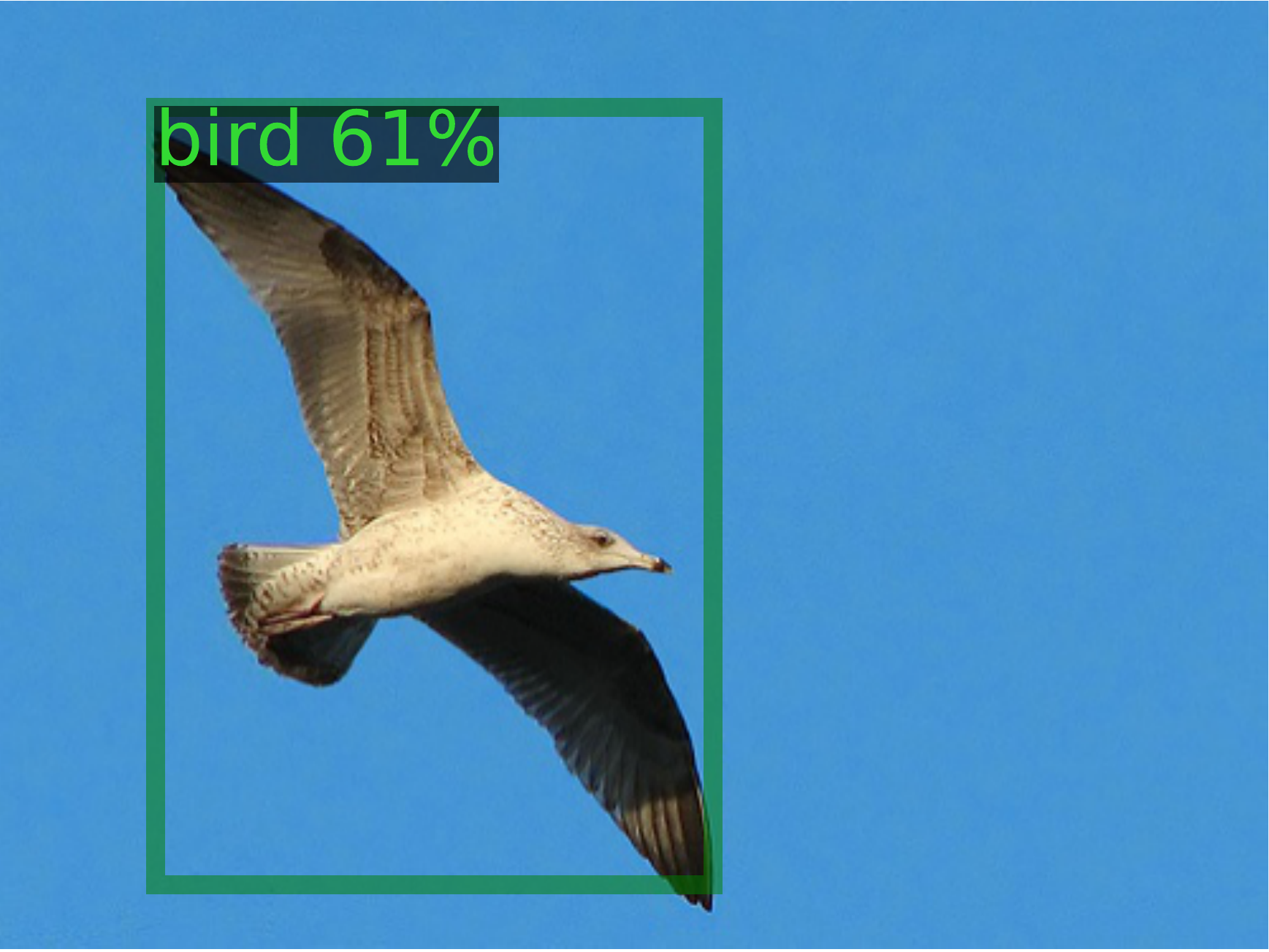} & \includegraphics[width=1in]{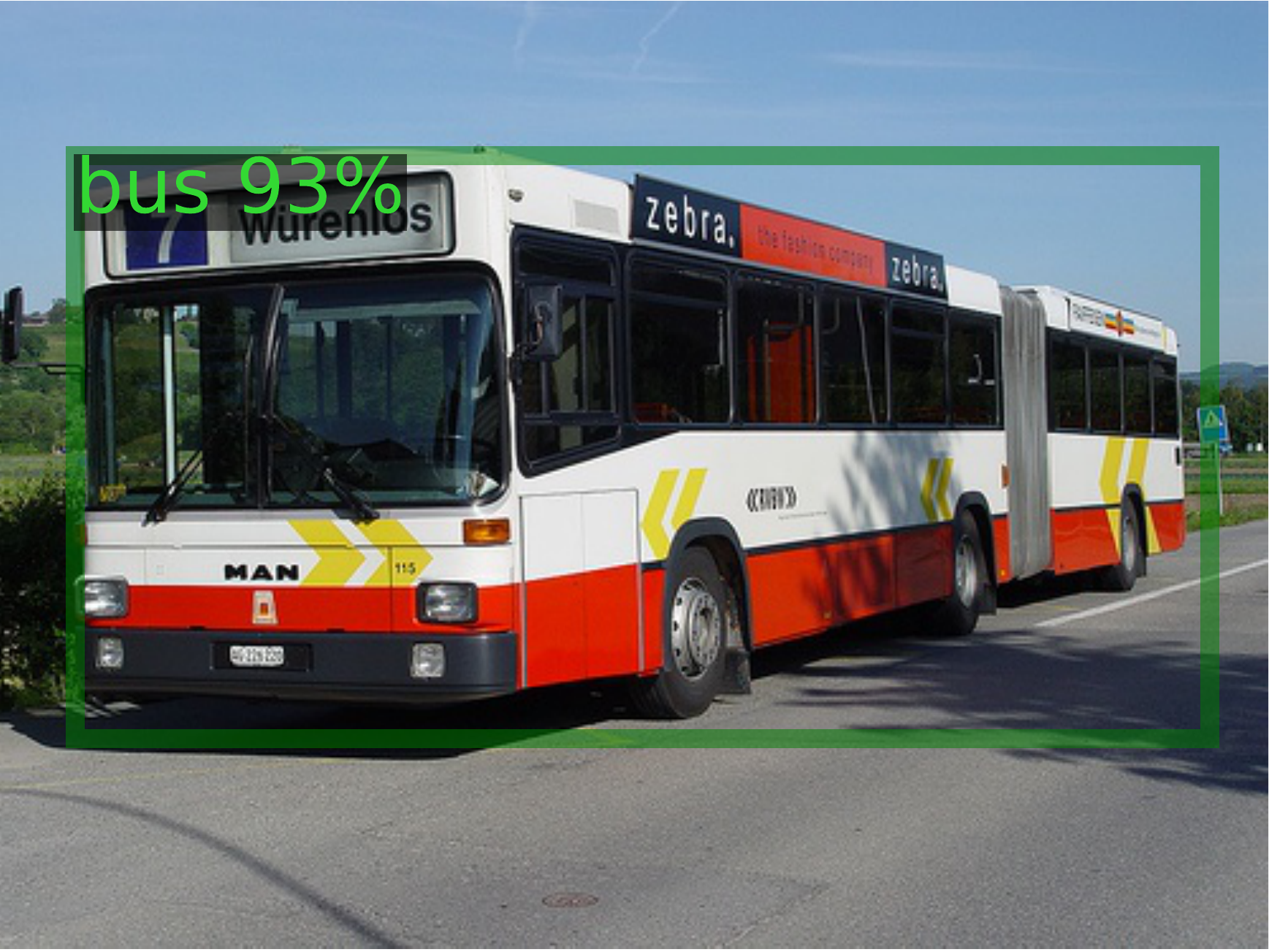} & \includegraphics[width=1in]{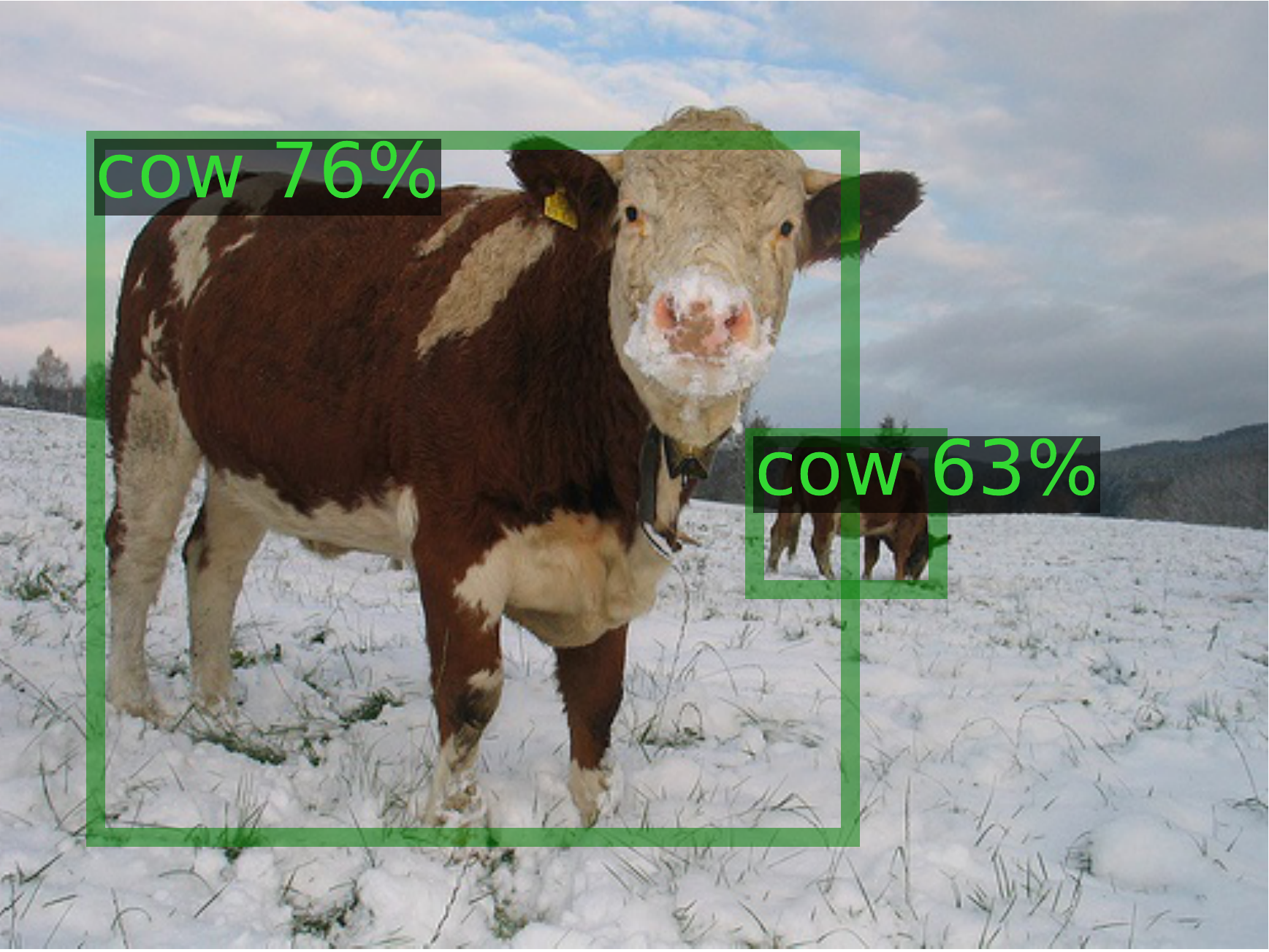} & \includegraphics[width=1in]{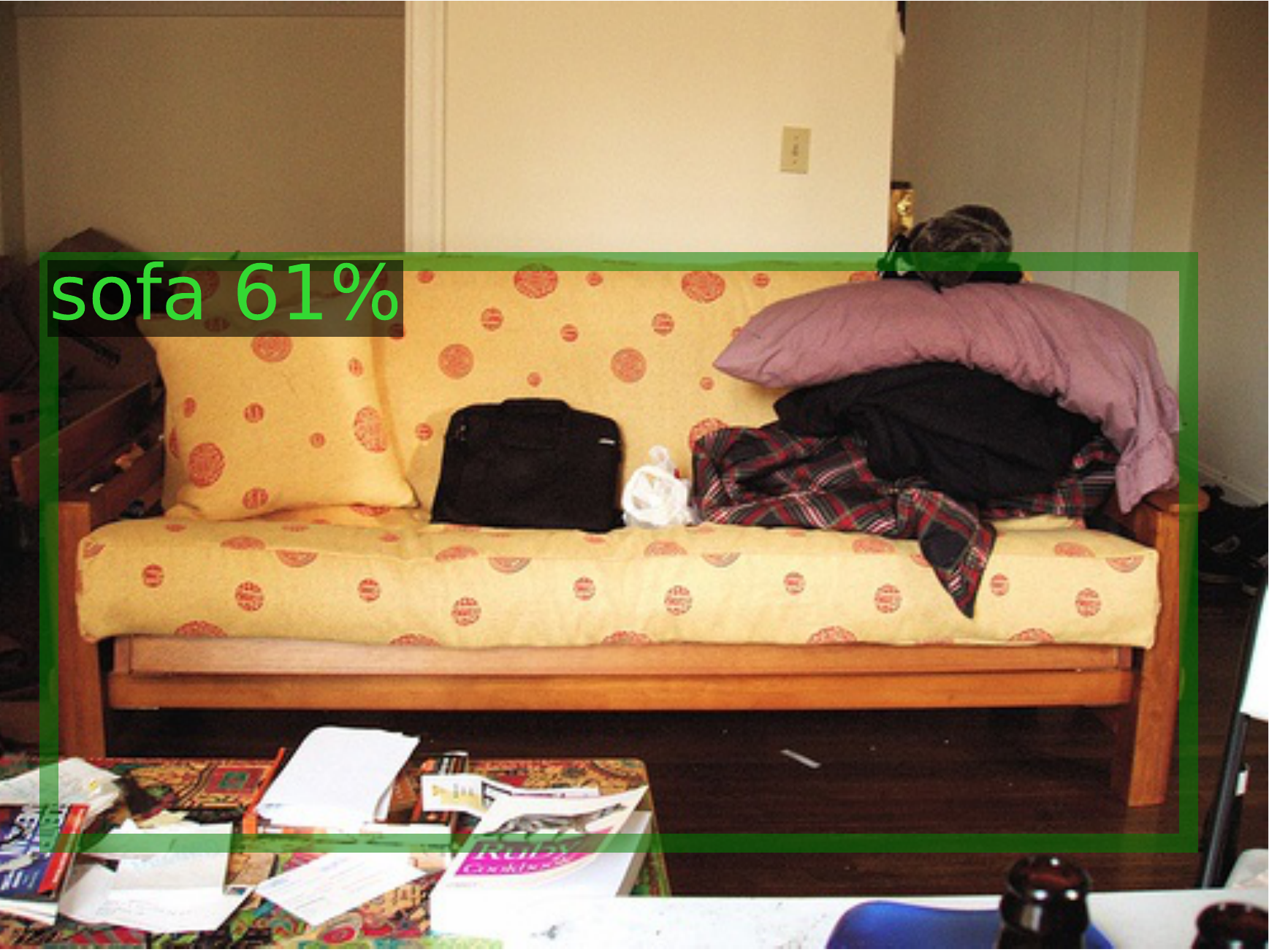} & \includegraphics[width=1in]{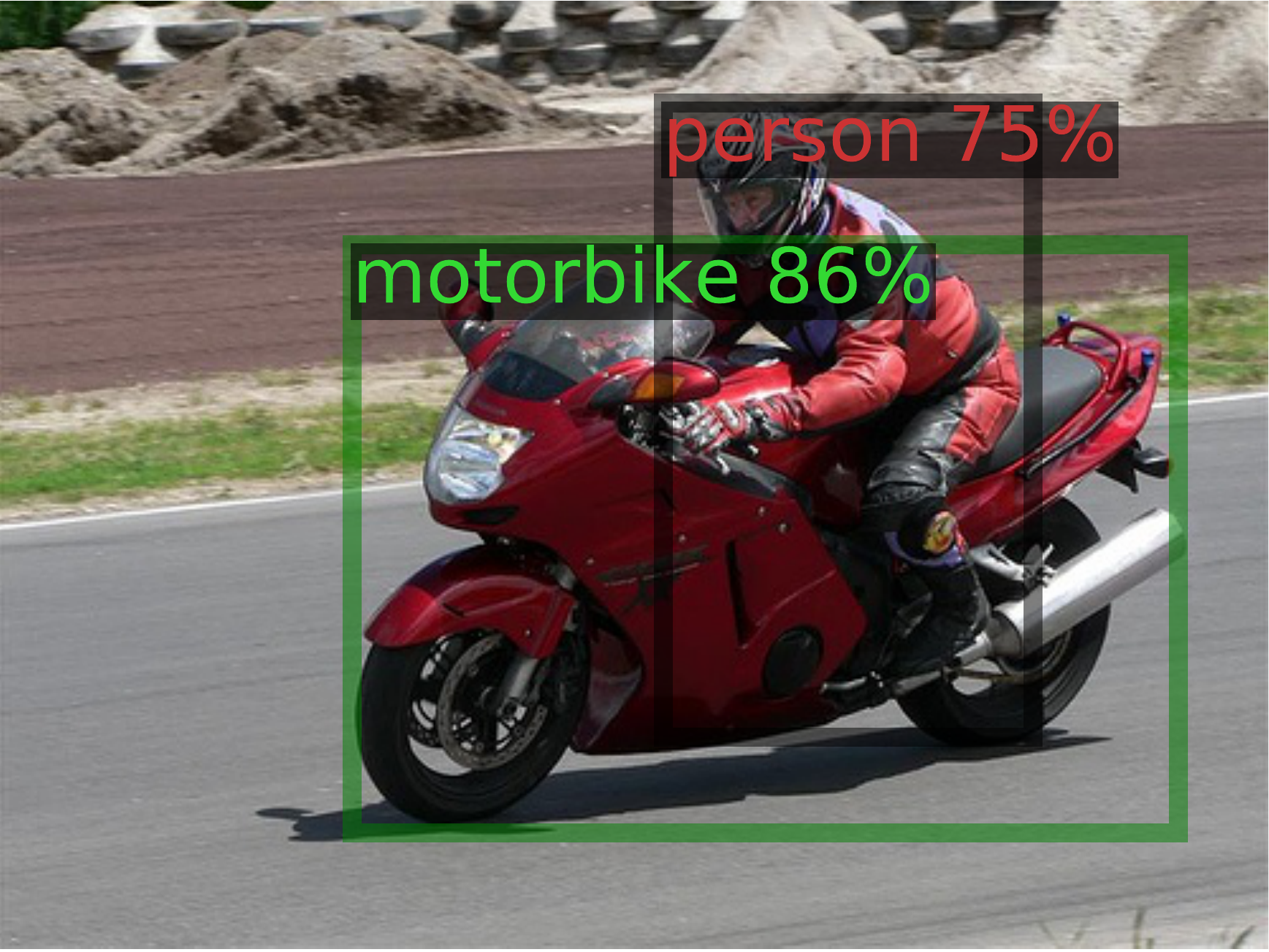} \\
			& \rotatebox{90}{\hspace{4mm}Failure} &
			\includegraphics[width=1in]{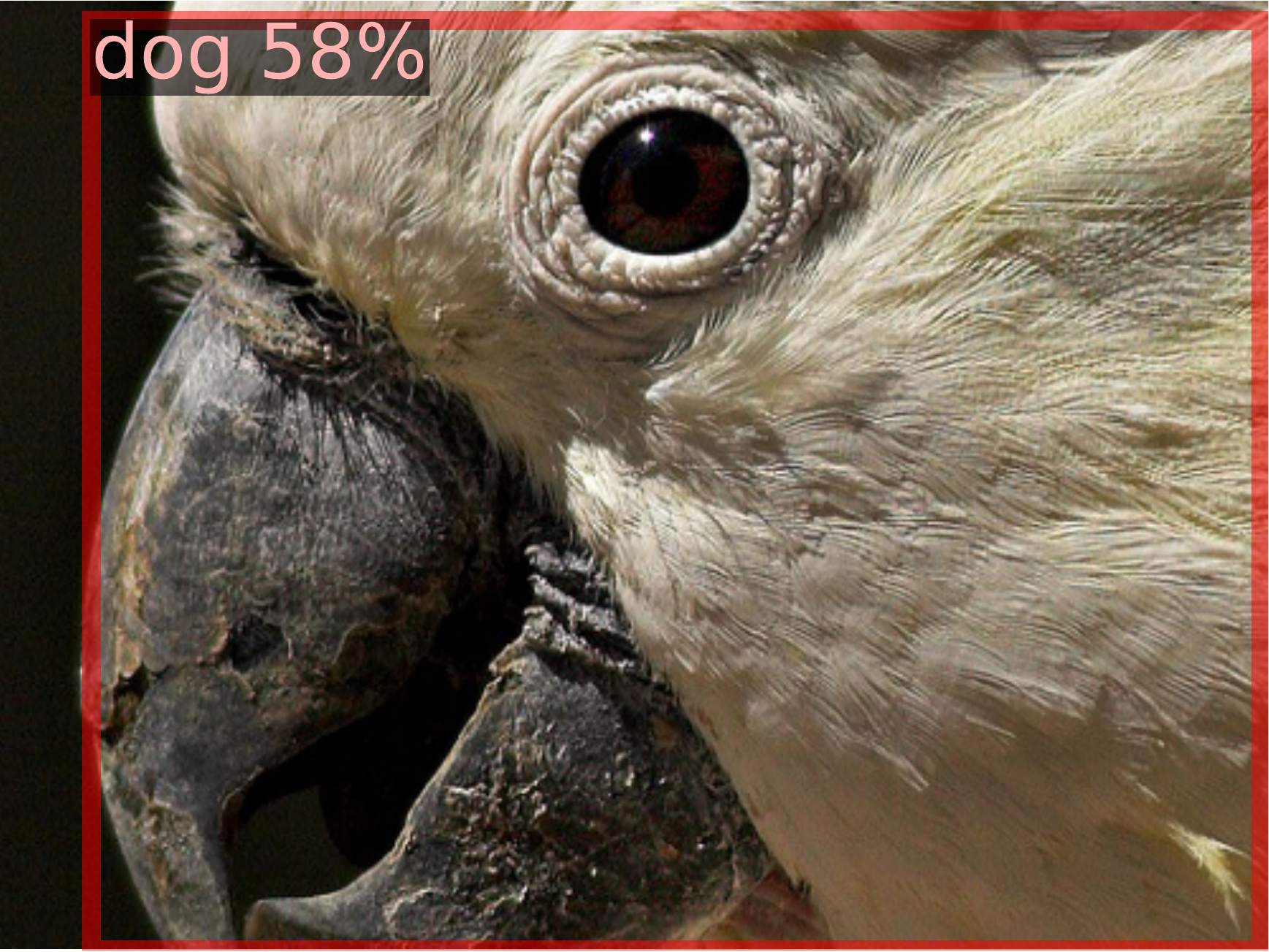} & \includegraphics[width=1in]{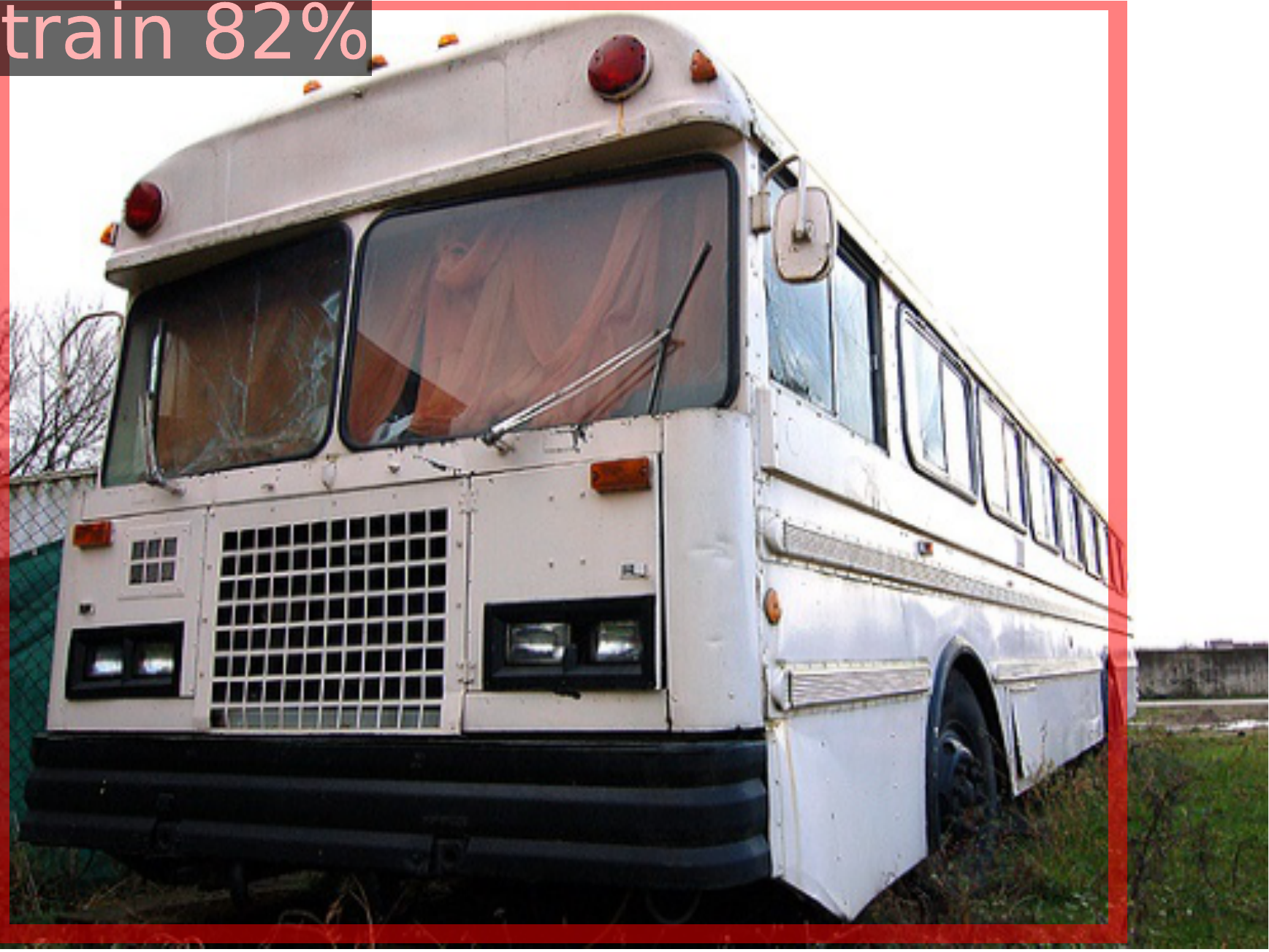} & \includegraphics[width=1in]{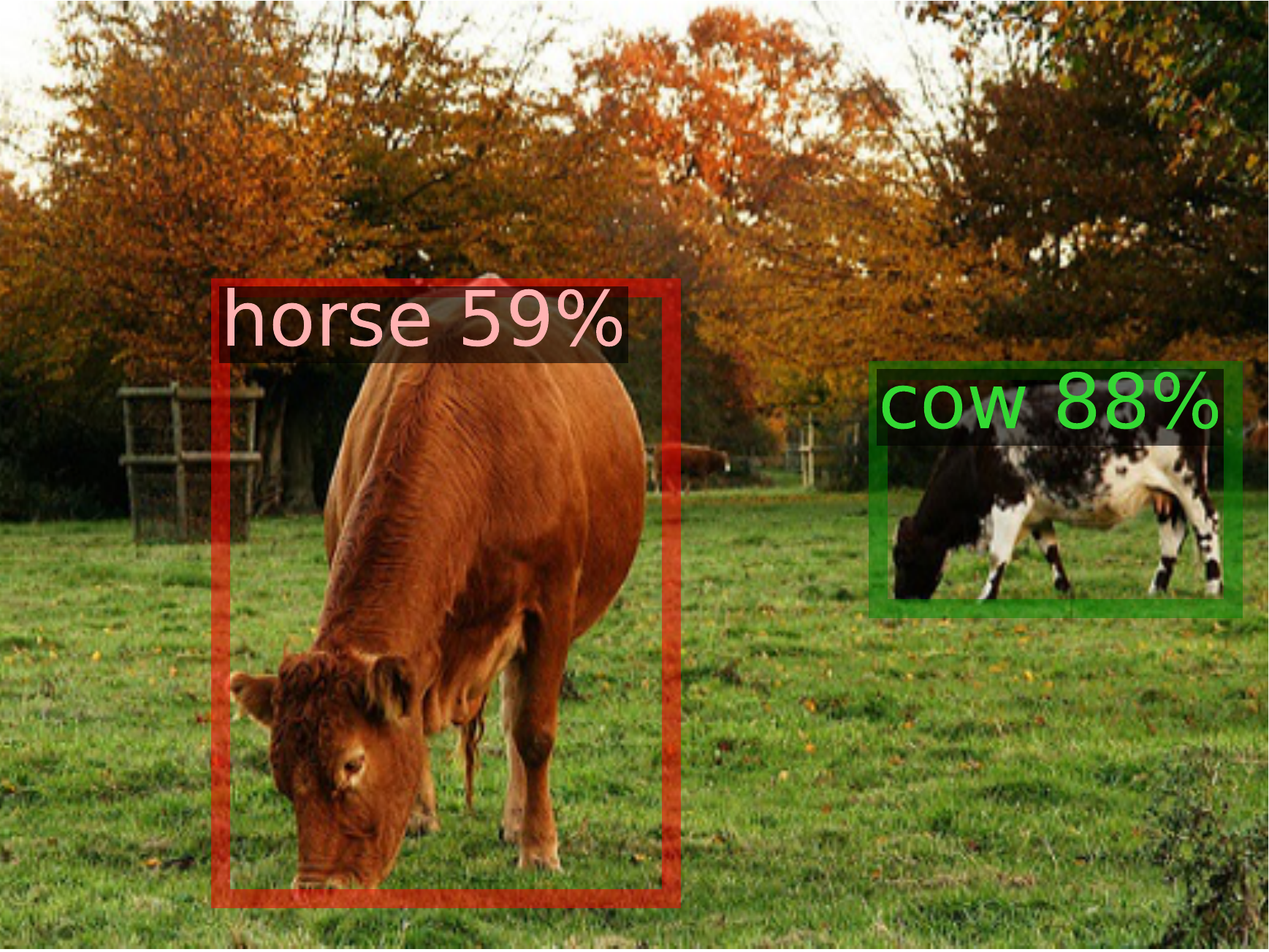} & \includegraphics[width=1in]{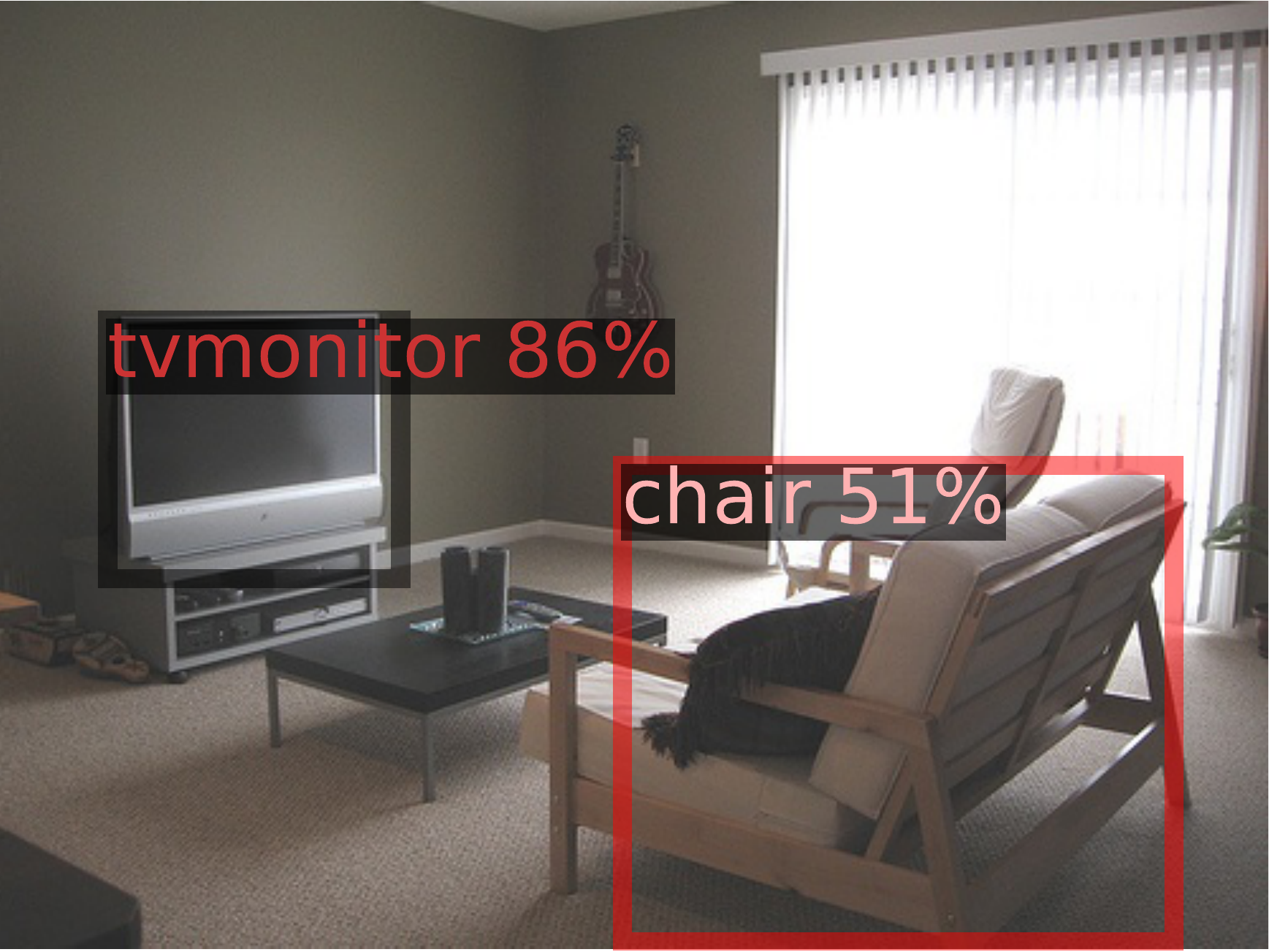} & \includegraphics[width=1in]{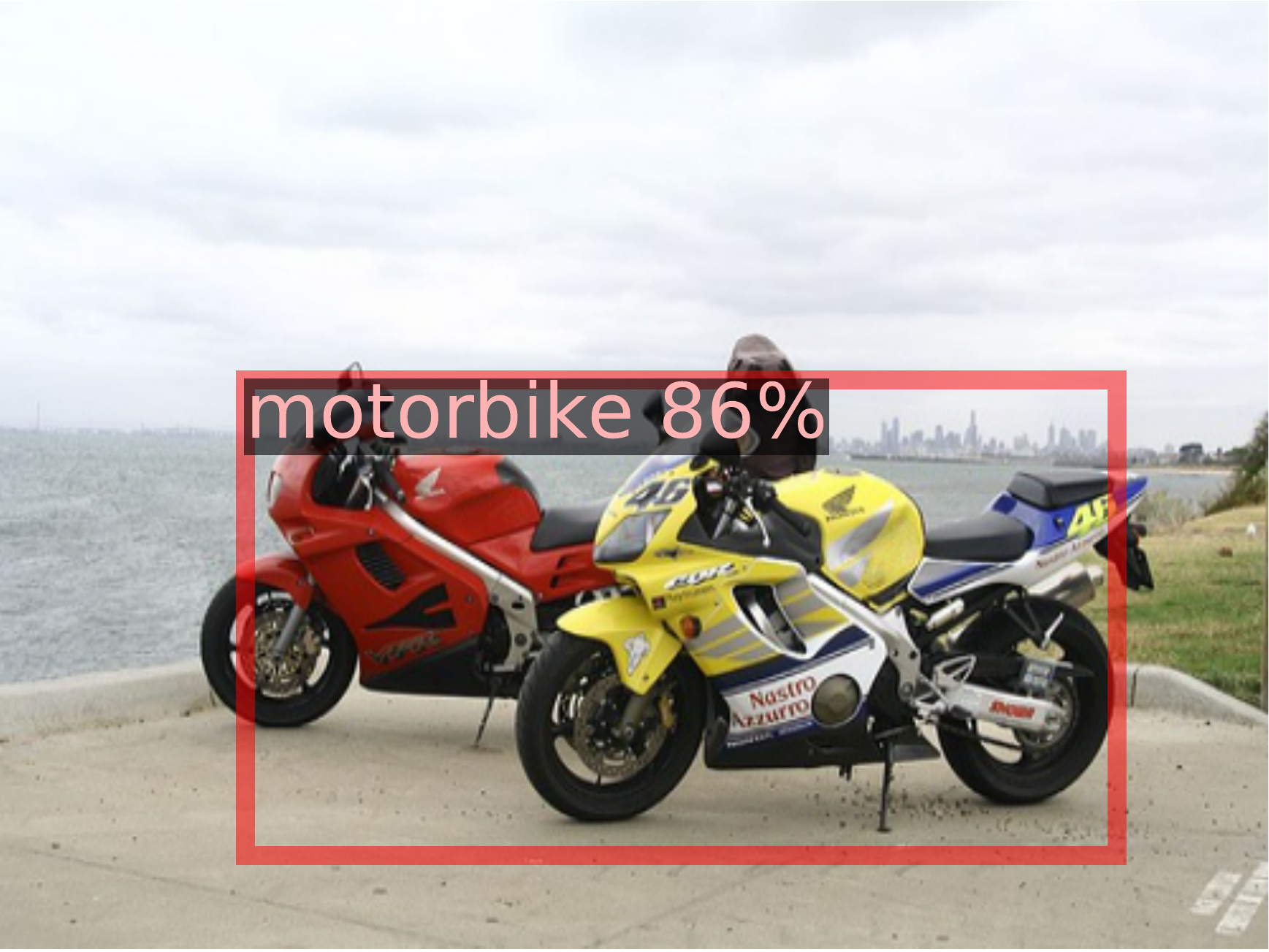} \\
			\multirow{2}{*}{\rotatebox{90}{COCO}} &
			\rotatebox{90}{\hspace{4mm}Success} &
			\includegraphics[width=1in]{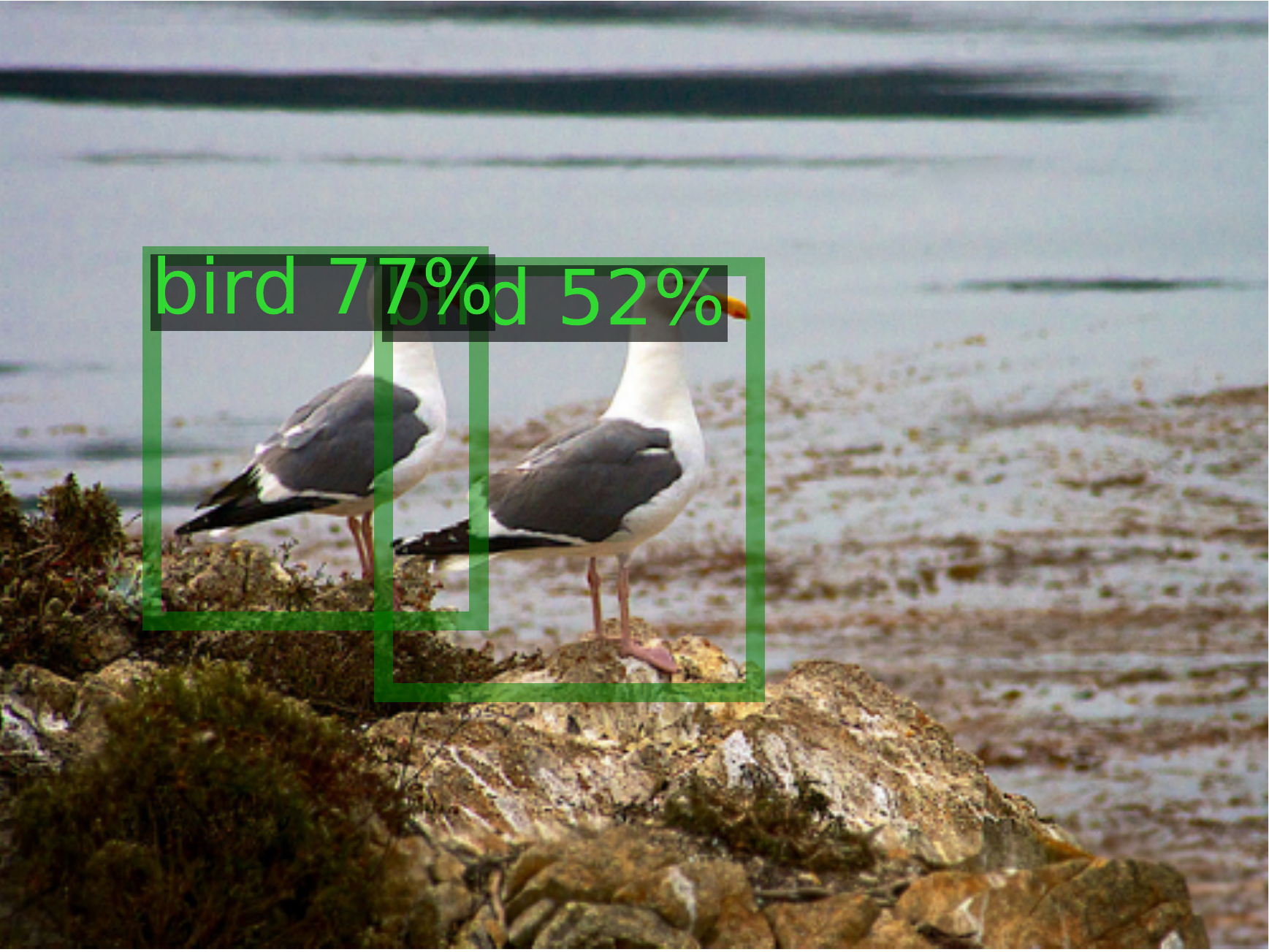} & \includegraphics[width=1in]{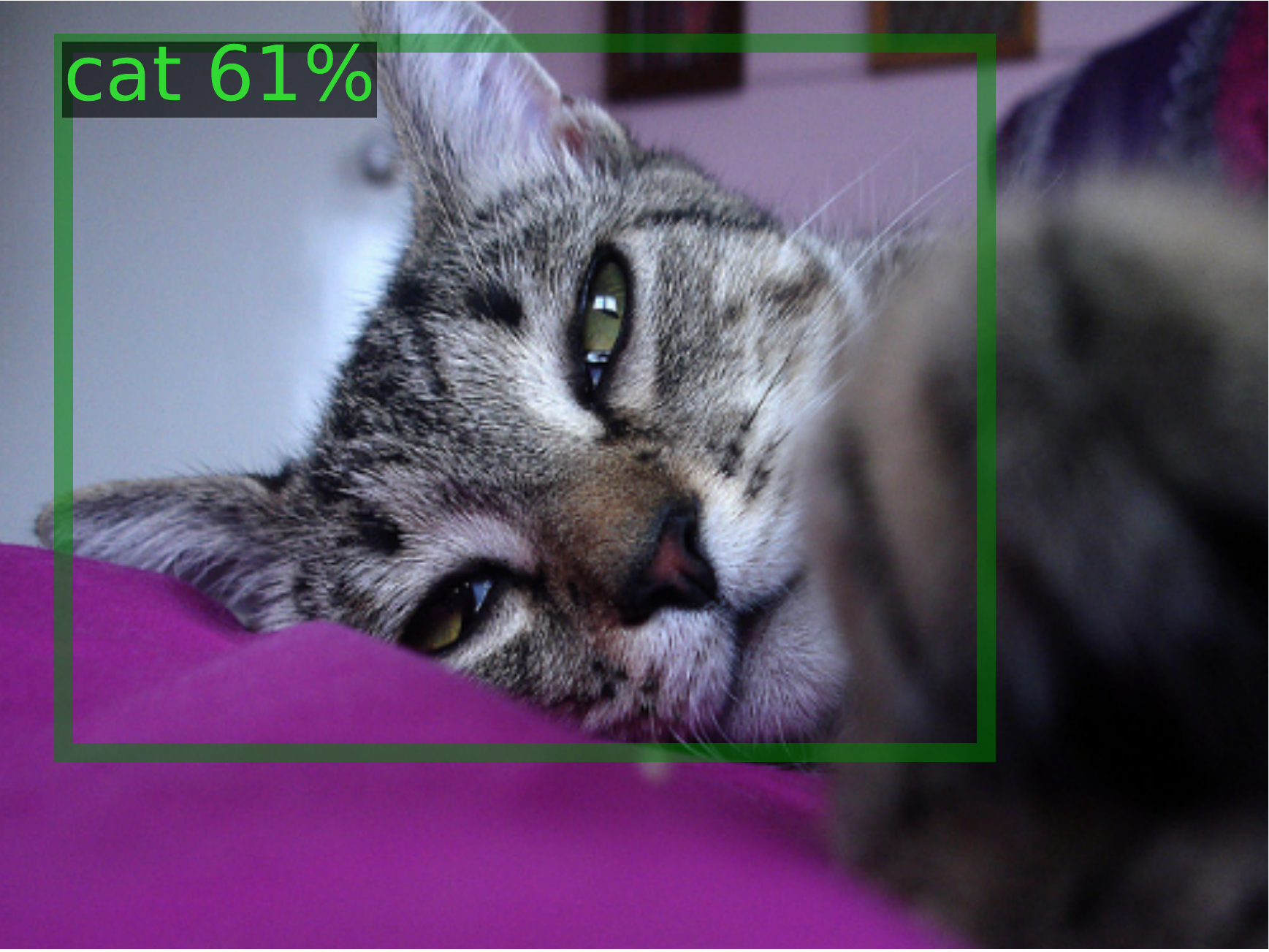} & \includegraphics[width=1in]{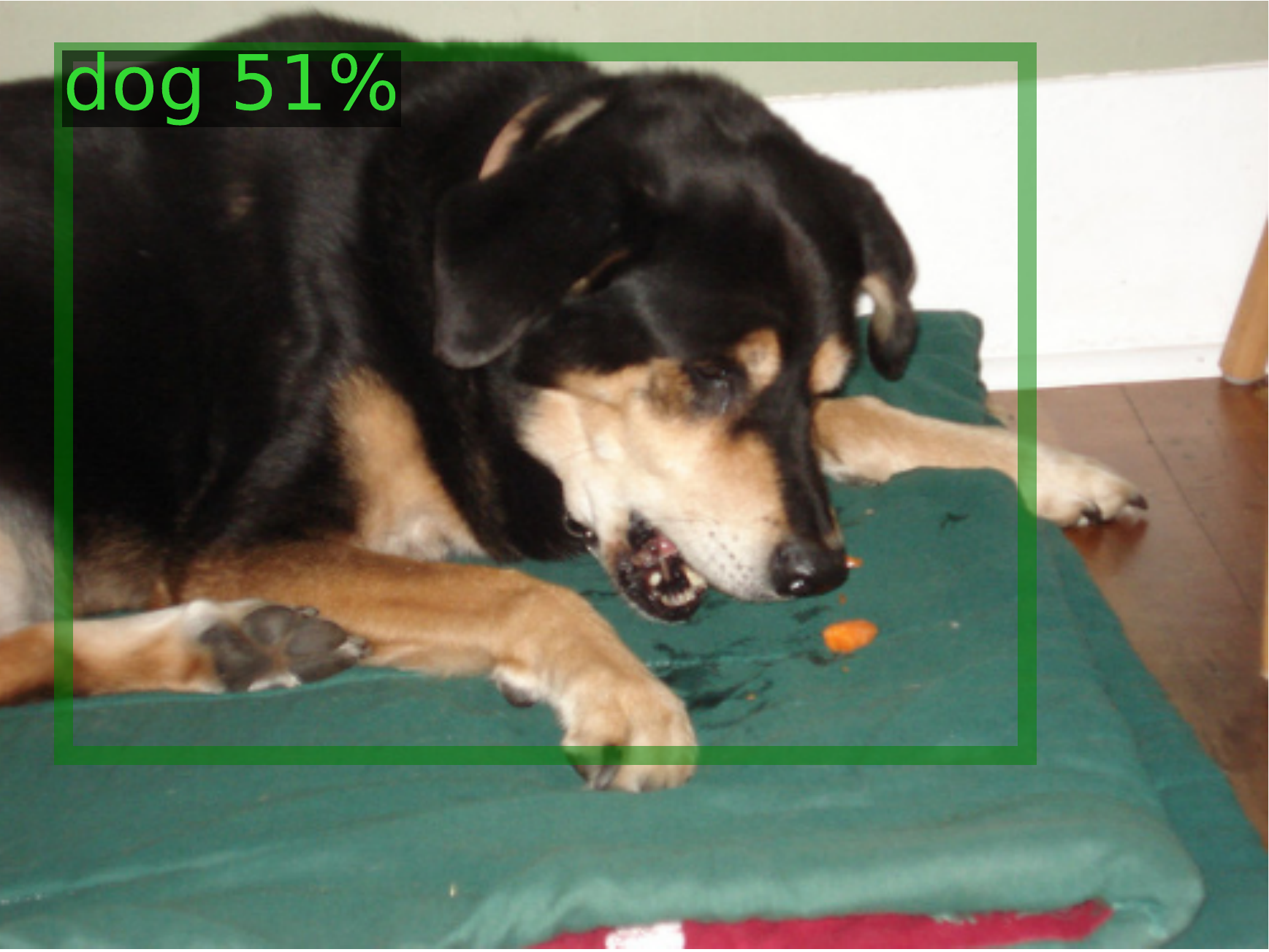} & \includegraphics[width=1in]{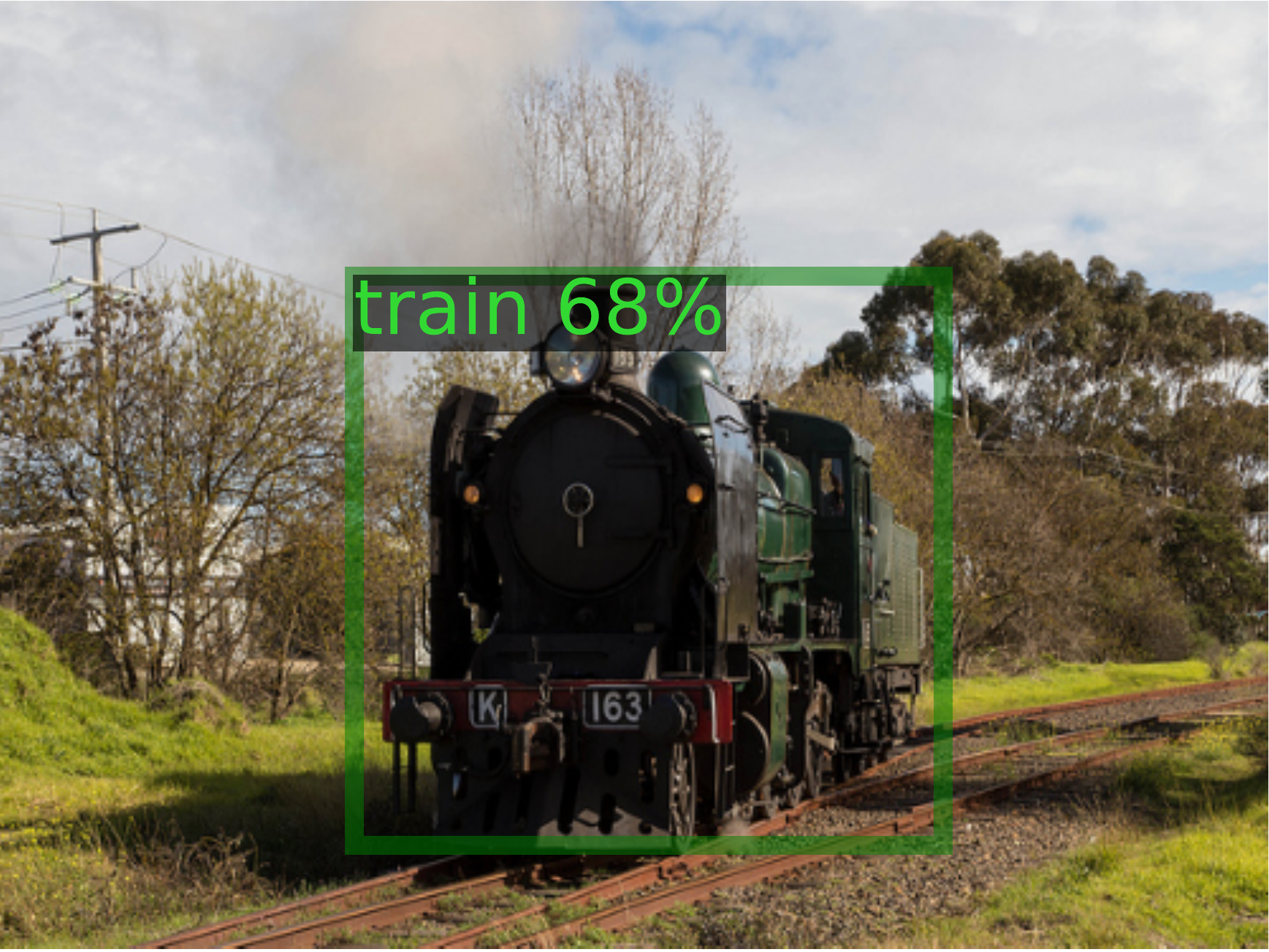} & \includegraphics[width=1in]{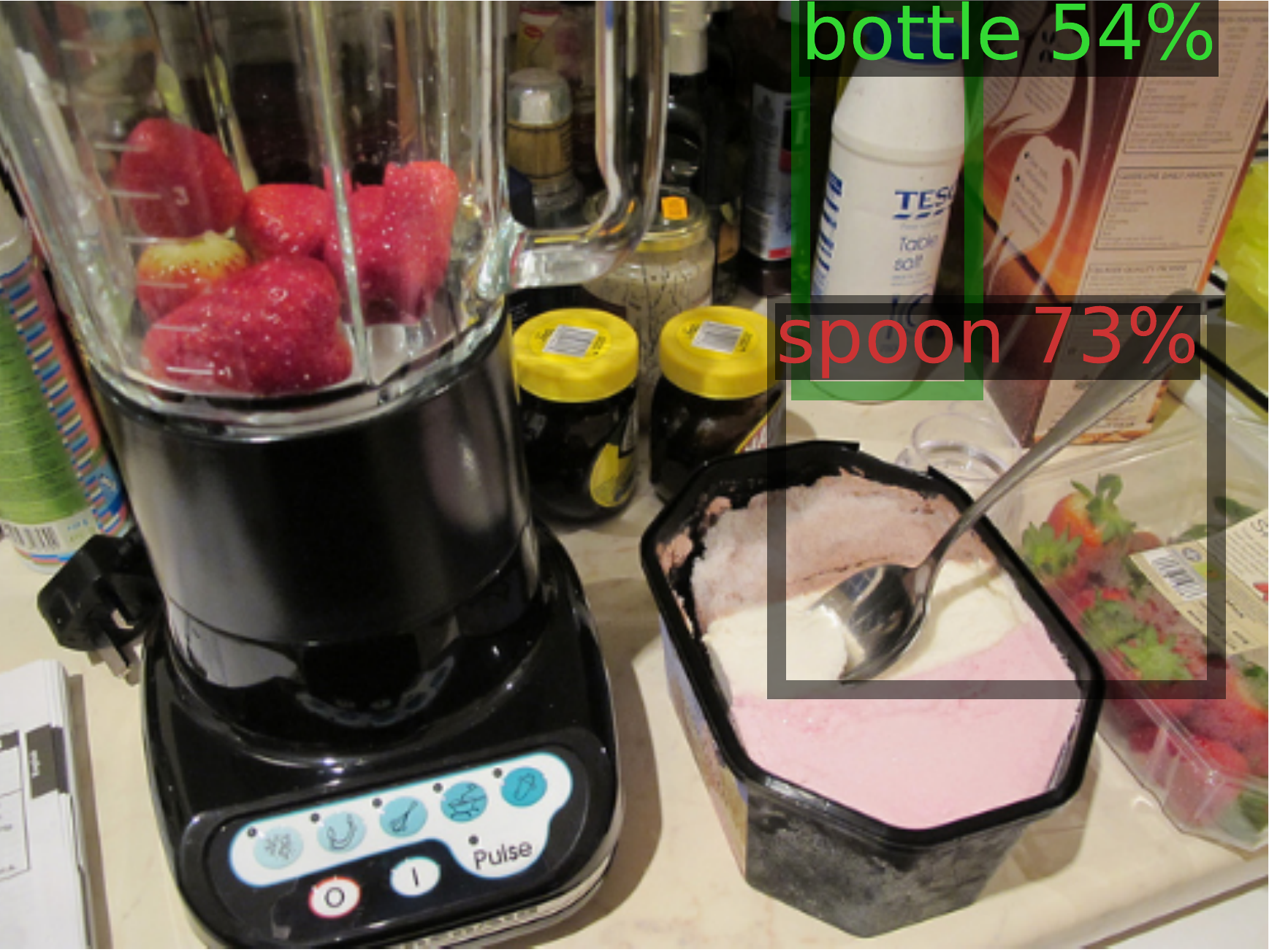} \\
			& \rotatebox{90}{\hspace{4mm}Failure} &
			\includegraphics[width=1in]{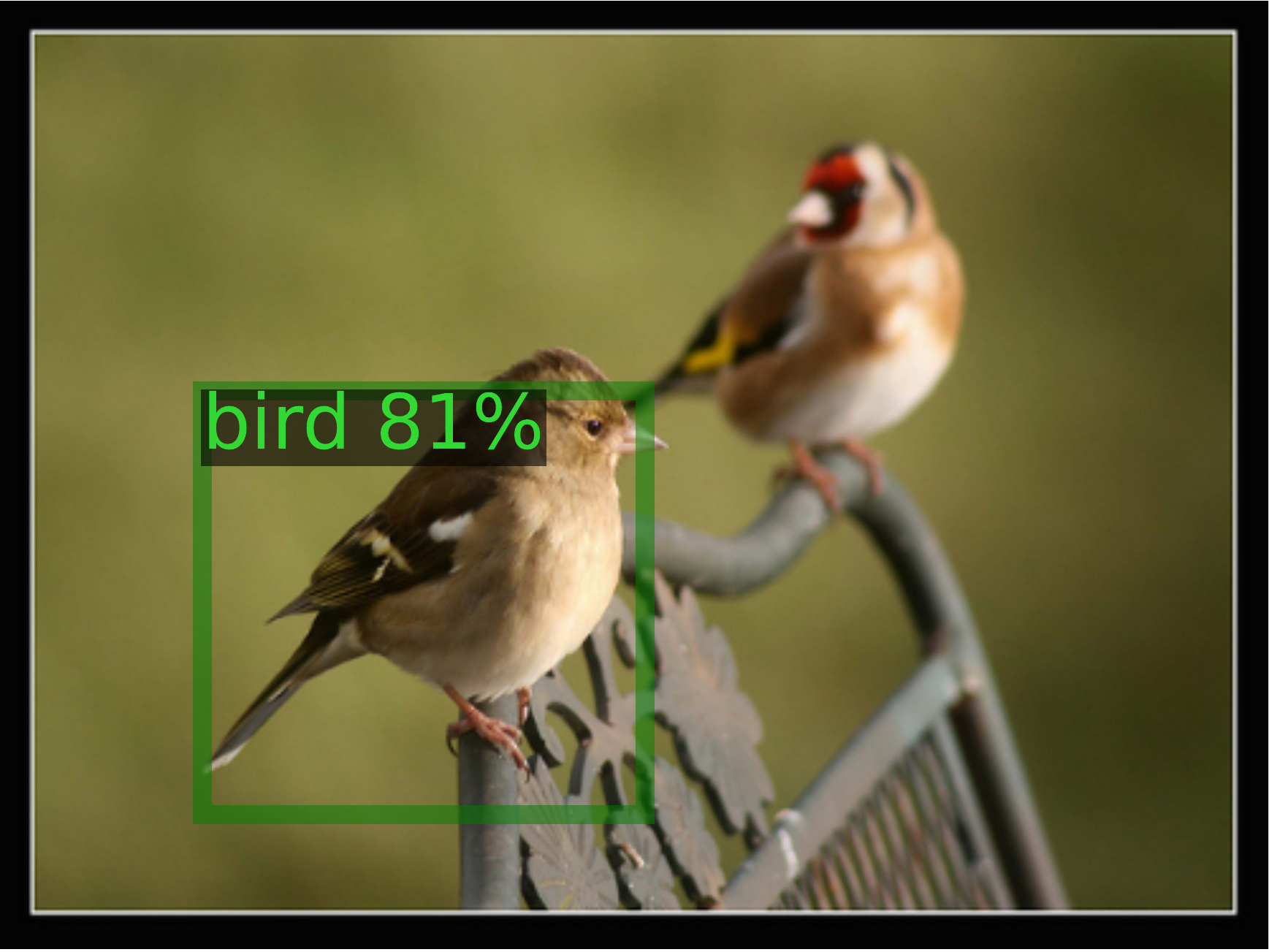} & \includegraphics[width=1in]{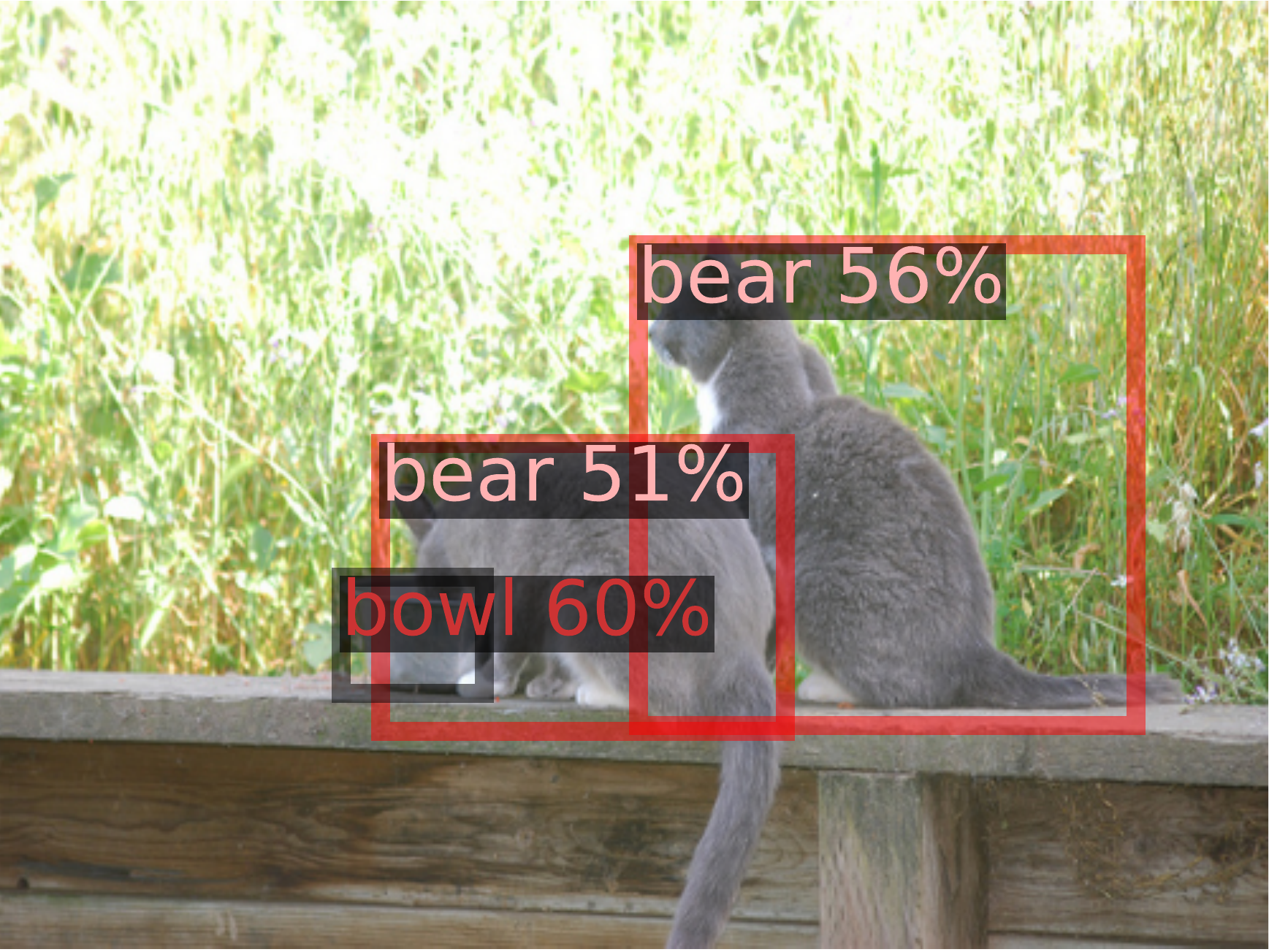} & \includegraphics[width=1in]{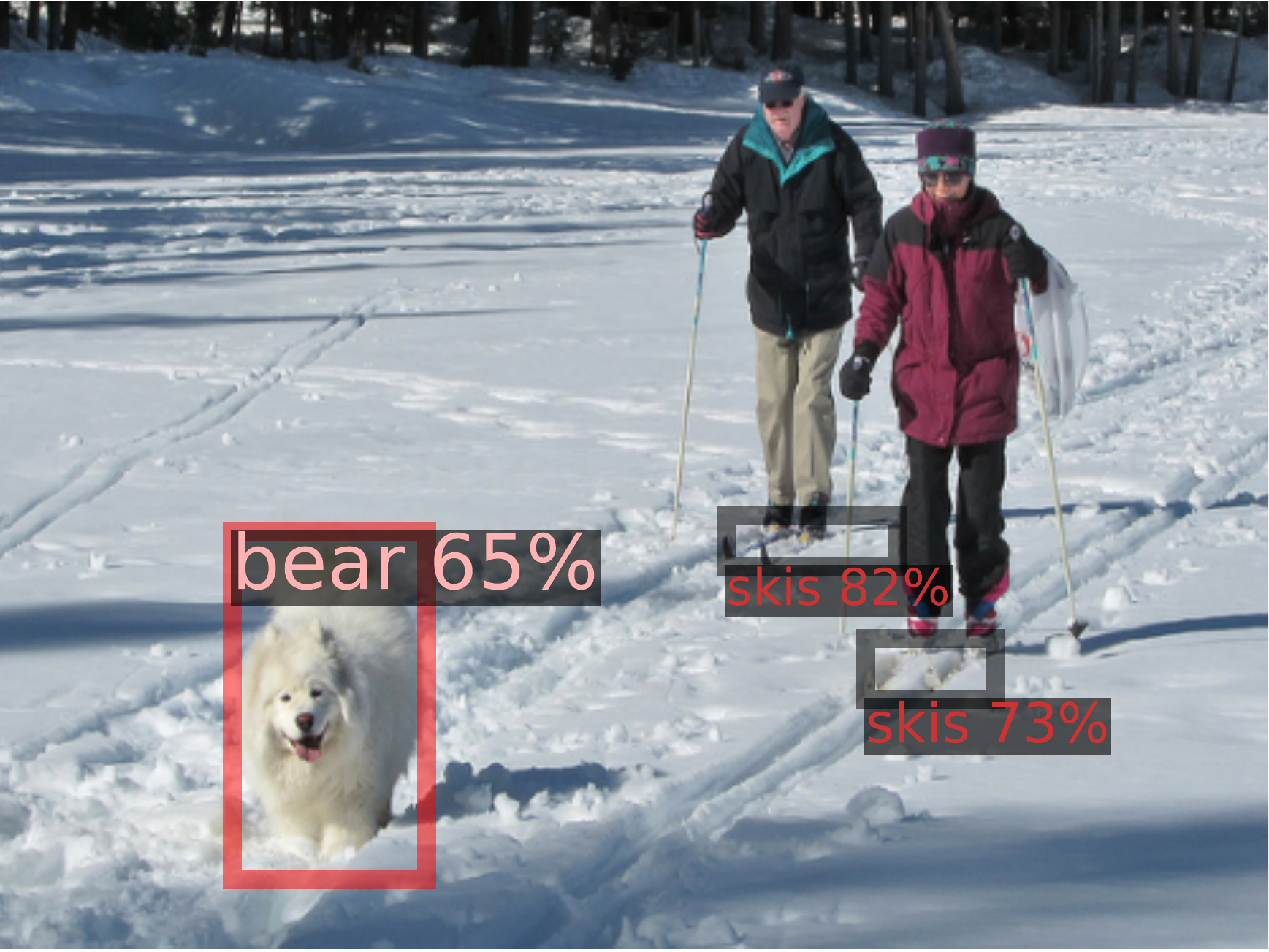} & \includegraphics[width=1in]{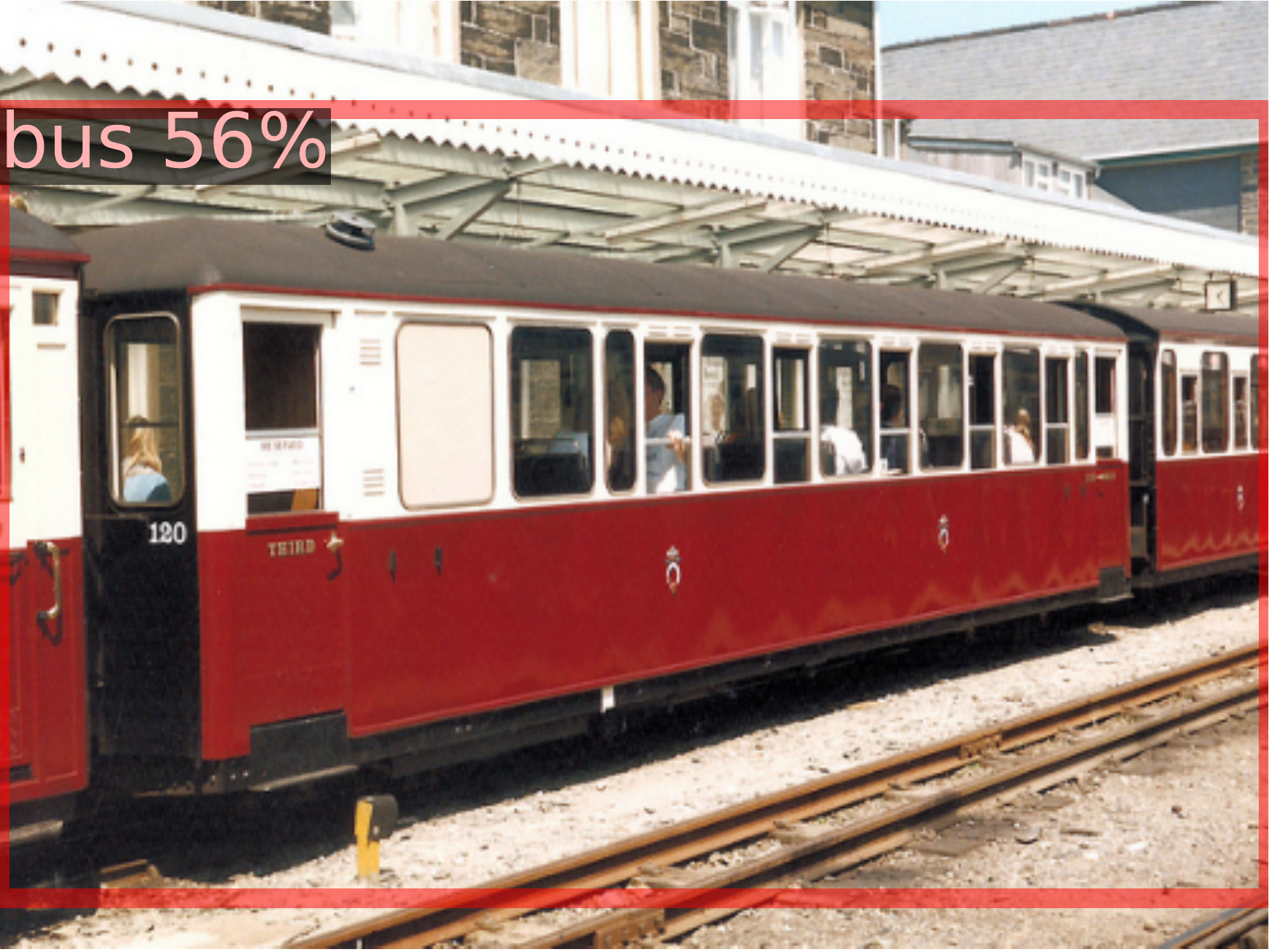} & \includegraphics[width=1in]{figs/coco_bottle_b.pdf} \\
	\end{tabular}}
    \caption{Success (green boxes) and failure (red boxes) cases of our approach on novel classes from split 1 of PASCAL VOC (bird, bus, cow, sofa, and motorbike) and COCO (bird, cat, dog, train, and bottle). The black boxes are detected objects of irrelevant classes, which can be ignored. \vspace{1mm}}
    \label{fig:det-vis}
\end{figure*}}

\subsection{Ablation study and visualization}
\label{sec:vis}

\minisection{Weight initialization.}
We explore two different ways of initializing the weights of the novel classifier before few-shot fine-tuning: (1) random initialization and (2) fine-tuning a predictor on the novel set and using the classifier's weights as initialization. We compare both methods on $K=1,3,10$ on split 3 of PASCAL VOC and COCO and show the results in Table~\ref{tab:weight_init}. 
On PASCAL VOC, simple random initialization can outperform initialization using fine-tuned novel weights. On COCO, using the novel weights can improve the performance over random initialization. This is probably due to the increased complexity and number of classes of COCO compared to PASCAL VOC. We use random initialization for all PASCAL VOC experiments and novel initialization for all COCO and LVIS experiments.

\minisection{Scaling factor of cosine similarity.}
We explore the effect of different scaling factors for computing cosine similarity. We compare three different factors, $\alpha=10,20,50$. We use the same evaluation setting as the previous ablation study and report the results in Table~\ref{tab:cos_scale}. On PASCAL VOC, $\alpha=20$ outperforms the other scale factors in both base AP and novel AP. On COCO, $\alpha=20$ achieves better novel AP at the cost of worse base AP. Since it has the best performance on novel classes across both datasets, we use $\alpha=20$ in all of our experiments with cosine similarity.

\begin{table}[!h]
	\centering
	\footnotesize
	\setlength{\tabcolsep}{0.4em}
	\caption{Ablation of weight initialization of the novel classifier. \vspace{2mm}}
	\adjustbox{width=.9\linewidth}{
		\begin{tabular}{ccccc|ccc}
			\multirow{2}{*}{Dataset}&\multirow{2}{*}{Init.}&\multicolumn{3}{c|}{Base AP} & \multicolumn{3}{c}{Novel AP}\\
			& & 1 & 3 & 10 & 1 & 3 & 10 \\ \midrule
			PASCAL VOC & Random & \textbf{51.2} & 52.6 & \textbf{52.8} & \textbf{15.6} & \textbf{25.0} & \textbf{29.4} \\ 
			(split 3) & Novel & 50.9 & \textbf{53.1} & 52.5 & 13.4 & 24.9 & 28.9 \\ \midrule
			\multirow{2}{*}{COCO} & Random & 34.0 & \textbf{34.7} & \textbf{34.6} & 3.2 & 6.4 & 9.6 \\ 
			& Novel & \textbf{34.1} & \textbf{34.7} & \textbf{34.6} & \textbf{3.4} & \textbf{6.6} & \textbf{9.8} \\
			\bottomrule
	\end{tabular}}
	\label{tab:weight_init} 
\end{table}

\begin{table}[!h]
	\centering
	\footnotesize
	\setlength{\tabcolsep}{0.4em}
	\caption{Ablation of scaling factor of cosine similarity. \vspace{1mm}}
	\adjustbox{width=.9\linewidth}{
		\begin{tabular}{ccccc|ccc}
			&&\multicolumn{3}{c|}{Base AP} & \multicolumn{3}{c}{Novel AP}\\
			Dataset & Scale  & 1 & 3 & 10 & 1 & 3 & 10 \\ \midrule
			 & 10 & 51.0 & \textbf{52.8} & 52.7 & 9.9 & 24.9 & 19.4 \\ 
			PASCAL VOC & 20 & \textbf{51.2} & 52.6 & \textbf{52.8} & \textbf{15.6} & \textbf{25.0} & \textbf{29.4} \\
			(split 3) & 50 & 47.4 & 48.7 & 50.4 & 13.2 & 22.5 & 27.6 \\ \midrule
			\multirow{3}{*}{COCO} & 10 & \textbf{34.3} & \textbf{34.9} & \textbf{35.0} & 2.8 & 3.4 & 4.7 \\ 
			& 20 & 34.1 & 34.7 & 33.9 & \textbf{3.4} & \textbf{6.6} & \textbf{10.0} \\
			& 50 & 30.1 & 30.7 & 34.3 & 2.4 & 5.4 & 9.0 \\
			\bottomrule
	\end{tabular}} 
	\label{tab:cos_scale}
\end{table}

\minisection{Detection results.}
We provide qualitative visualizations of the detected novel objects on PASCAL VOC and COCO in Figure~\ref{fig:det-vis}.
We show both success (green boxes) and failure cases (red boxes) when detecting novel objects for each dataset to help analyze the possible error types.
On the first split of PASCAL VOC, we visualize the results of our 10-shot \model w/ cos model.
On COCO, we visualize the results of the 30-shot \texttt{\model w/cos} model.
The failure cases include misclassifying novel objects as similar base objects, \textit{e.g.}, row 2 columns 1, 2, 3, and 4, mislocalizing the objects, \textit{e.g.}, row 2 column 5, and missing detections, \textit{e.g.}, row 4 columns 1 and 5.



\section{Conclusion}
We proposed a simple two-stage fine-tuning approach for few-shot object detection.
Our method outperformed the previous meta-learning methods by a large margin on the current benchmarks.
In addition, we built more reliable benchmarks with revised evaluation protocols.
On the new benchmarks, our models achieved new states of the arts, and on the LVIS dataset our models improved the AP of rare classes by 4 points with negligible reduction of the AP of frequent classes.

\subsubsection*{Acknowledgments}
This work was supported by Berkeley AI Research, RISE Lab, Berkeley DeepDrive and DARPA. 

	
\bibliographystyle{icml2020}
\bibliography{references}

\appendix
\section{Generalized Object Detection Benchmarks}
We present the full benchmark results of PASCAL VOC (Table~\ref{tab:voc_bench}) and COCO
(Table~\ref{tab:coco_bench}) on the revised benchmark used in this work. We report the average AP, AP50 and AP75 for all the classes, base classes only, and novel classes only in the tables. For each evaluation metric, we report the average value of $n$ repeated runs with different groups of randomly sampled training shots (30 for PASCAL VOC and 10 for COCO) as well as the 95\% confidence interval estimate of the mean values.  The 95\% confidence interval is calculated by 
\begin{equation}
    95\% \; CI = 1.96 \cdot \frac{s}{\sqrt{n}},
\end{equation}
where 1.96 is the $Z$-value, $s$ is the standard deviation, and $n$ is the number of repeated runs.

We compare two of our methods, one using a FC-based classifier (\texttt{\model w/fc}) and one using a cosine similarity based classifier (\texttt{\model w/cos}). We also compare against a fine-tuning baseline \texttt{FRCN+ft-full} and against FSRW~\cite{kang2019few} using their released code on PASCAL VOC shown in Table~\ref{tab:voc_bench}. 

As shown in Table~\ref{tab:voc_bench},
\texttt{\model w/cos} is able to significantly outperform \texttt{\model w/fc} in overall AP across most splits and shots.
We observe that using a cosine similarity based classifier can achieve much higher accuracy on base classes, especially in higher shots.
On split 1 and 3, \texttt{\model w/cos} is able to outperform \texttt{\model w/fc} by over 3 points on bAP on 5 and 10 shots.
Across all shots in split 1, \texttt{\model w/cos} consistently outperforms \texttt{\model w/fc} on nAP75 by over 2 points in the novel classes. 

Moreover, the AP of our models is usually over 10 points higher than that of \texttt{FRCN+ft-full} and FSRW on all settings. Note that FSRW uses YOLOv2 as the base object detector, while we are using Faster R-CNN. ~\citet{wang2019meta} shows that there are only about 2 points of difference when using a one or two-stage detector. Therefore, our improvements should still be significant despite the difference in the base detector.

We evaluate on COCO over six different number of shots $K=1,2,3,5,10,30$ shown in Table~\ref{tab:coco_bench}.
Although the differences are less significant than on PASCAL VOC, similar observations can be made about accuracy on base classes and novel classes.

\begin{table*}[!h]
\centering
\footnotesize
\setlength{\tabcolsep}{0.4em}
\caption{Generalized object detection benchmarks on PASCAL VOC. For each metric, we report the average and 95\% confidence interval computed over 30 random samples. \vspace{1mm}}
\adjustbox{width=\linewidth}{
\begin{tabular}{c|c|c|ccc|ccc|ccc}
\toprule
\multirow{2}{*}{Split} & \multirow{2}{*}{\# shots} &\multirow{2}{*}{Method} &  \multicolumn{3}{c|}{Overall}  &\multicolumn{3}{c|}{Base class} & \multicolumn{3}{c}{Novel class} \\ \cmidrule{4-12}
& & & AP & AP50 & AP75 & bAP & bAP50 & bAP75 & nAP & nAP50 & nAP75 \\ \midrule
\multirow{21}{*}{Split 1} & 
\multirow{4}{*}{1}
    & FSRW~\cite{kang2019few} & 27.6 $ \pm $ 0.5 & 50.8 $ \pm $ 0.9 & 26.5 $ \pm $ 0.6 & 34.1 $ \pm $ 0.5 & 62.9 $ \pm $ 0.9 & 32.6 $ \pm $ 0.5 & 8.0 $ \pm $ 1.0 & 14.2 $ \pm $ 1.7 & 7.9 $ \pm $ 1.1\\
    & & FRCN+ft-full & 30.2$\pm$0.6 & 49.4$\pm$0.7 & 32.2$\pm$0.9 & 38.2$\pm$0.8 & 62.6$\pm$1.0 & 40.8$\pm$1.1 & 6.0$\pm$0.7 & 9.9$\pm$1.2 & 6.3$\pm$0.8\\
     & & {\model w/fc} & 39.6$\pm$0.5 & 63.5$\pm$0.7 & 43.2$\pm$0.7 & 48.7$\pm$0.7 & 77.1$\pm$0.7 & 53.7$\pm$1.0 & 12.2$\pm$1.6 & 22.9$\pm$2.5 & 11.6$\pm$1.9 \\
    & &\cellcolor{Gray} {\model w/cos} & \cellcolor{Gray}40.6$\pm$0.5 & \cellcolor{Gray}64.5$\pm$0.6 & \cellcolor{Gray}44.7$\pm$0.6 & \cellcolor{Gray}49.4$\pm$0.4 & \cellcolor{Gray}77.6$\pm$0.2 & \cellcolor{Gray}54.8$\pm$0.5 & \cellcolor{Gray}14.2$\pm$1.4 &\cellcolor{Gray} 25.3$\pm$2.2 & \cellcolor{Gray}14.2$\pm$1.8 \\ \cmidrule{2-12}
    & \multirow{4}{*}{2} & FSRW~\cite{kang2019few} & 
    28.7$\pm$0.4&52.2$\pm$0.6&27.7$\pm$0.5&33.9$\pm$0.4&61.8$\pm$0.5&32.7$\pm$0.5&13.2$\pm$1.0&23.6$\pm$1.7&12.7$\pm$1.1 \\
    & & FRCN+ft-full & 30.5$\pm$0.6 & 49.4$\pm$0.8 & 32.6$\pm$0.7 & 37.3$\pm$0.7 & 60.7$\pm$1.0 & 40.1$\pm$0.9 & 9.9$\pm$0.9 & 15.6$\pm$1.4 & 10.3$\pm$1.0 \\
    & & {\model w/fc} & 40.5$\pm$0.5 & 65.5$\pm$0.7 & 43.8$\pm$0.7 & 47.8$\pm$0.7 & 75.8$\pm$0.7 & 52.2$\pm$1.0 & 18.9$\pm$1.5 & 34.5$\pm$2.4 & 18.4$\pm$1.9  \\
    & &\cellcolor{Gray} {\model w/cos} & \cellcolor{Gray}42.6$\pm$0.3 & \cellcolor{Gray} 67.1$\pm$0.4 & \cellcolor{Gray}47.0$\pm$0.4 &\cellcolor{Gray} 49.6$\pm$0.3 & \cellcolor{Gray}77.3$\pm$0.2 & \cellcolor{Gray}55.0$\pm$0.4 & \cellcolor{Gray}21.7$\pm$1.0 & \cellcolor{Gray}36.4$\pm$1.6 & \cellcolor{Gray}22.8$\pm$1.3  \\ \cmidrule{2-12}
    & \multirow{4}{*}{3} & FSRW~\cite{kang2019few} & 29.5$\pm$0.3&53.3$\pm$0.6&28.6$\pm$0.4&33.8$\pm$0.3&61.2$\pm$0.6&32.7$\pm$0.4&16.8$\pm$0.9&29.8$\pm$1.6&16.5$\pm$1.0 \\
    & & FRCN+ft-full & 31.8$\pm$0.5 & 51.4$\pm$0.8 & 34.2$\pm$0.6 & 37.9$\pm$0.5 & 61.3$\pm$0.7 & 40.7$\pm$0.6 & 13.7$\pm$1.0 & 21.6$\pm$1.6 & 14.8$\pm$1.1 \\
    & & {\model w/fc} & 41.8$\pm$0.9 & 67.1$\pm$0.9 & 45.4$\pm$1.2 & 48.2$\pm$0.9 & 76.0$\pm$0.9 & 53.1$\pm$1.2 & 22.6$\pm$1.2 & 40.4$\pm$1.7 & 22.4$\pm$1.7  \\
    & & \cellcolor{Gray}{\model w/cos} &\cellcolor{Gray} 43.7$\pm$0.3 &\cellcolor{Gray} 68.5$\pm$0.4 & \cellcolor{Gray}48.3$\pm$0.4 & \cellcolor{Gray}49.8$\pm$0.3 & \cellcolor{Gray}77.3$\pm$0.2 & \cellcolor{Gray}55.4$\pm$0.4 & \cellcolor{Gray}25.4$\pm$0.9 & \cellcolor{Gray}42.1$\pm$1.5 & \cellcolor{Gray}27.0$\pm$1.2  \\ \cmidrule{2-12}
    & \multirow{4}{*}{5} & FSRW~\cite{kang2019few} & 30.4$\pm$0.3&54.6$\pm$0.5&29.6$\pm$0.4&33.7$\pm$0.3&60.7$\pm$0.4&32.8$\pm$0.4&20.6$\pm$0.8&36.5$\pm$1.4&20.0$\pm$0.9 \\
    & & FRCN+ft-full & 32.7$\pm$0.5 & 52.5$\pm$0.8 & 35.0$\pm$0.6 & 37.6$\pm$0.4 & 60.6$\pm$0.6 & 40.3$\pm$0.5 & 17.9$\pm$1.1 & 28.0$\pm$1.7 & 19.2$\pm$1.3 \\
    & & {\model w/fc} & 41.9$\pm$0.6 & 68.0$\pm$0.7 & 45.0$\pm$0.8 & 47.2$\pm$0.6 & 75.1$\pm$0.6 & 51.5$\pm$0.8 & 25.9$\pm$1.0 & 46.7$\pm$1.4 & 25.3$\pm$1.2  \\
    & &\cellcolor{Gray} {\model w/cos} &\cellcolor{Gray} 44.8$\pm$0.3 & \cellcolor{Gray}70.1$\pm$0.4 &\cellcolor{Gray} 49.4$\pm$0.4 &\cellcolor{Gray} 50.1$\pm$0.2 &\cellcolor{Gray} 77.4$\pm$0.3 &\cellcolor{Gray} 55.6$\pm$0.3 &\cellcolor{Gray} 28.9$\pm$0.8 & \cellcolor{Gray}47.9$\pm$1.2 & \cellcolor{Gray}30.6$\pm$1.0  \\ \cmidrule{2-12}
    & \multirow{3}{*}{10} & FRCN+ft-full & 33.3$\pm$0.4 & 53.8$\pm$0.6 & 35.5$\pm$0.4 & 36.8$\pm$0.4 & 59.8$\pm$0.6 & 39.2$\pm$0.4 & 22.7$\pm$0.9 & 35.6$\pm$1.5 & 24.4$\pm$1.0 \\
    & & {\model w/fc} & 42.8$\pm$0.3 & 69.5$\pm$0.4 & 46.0$\pm$0.4 & 47.3$\pm$0.3 & 75.4$\pm$0.3 & 51.6$\pm$0.4 & 29.3$\pm$0.7 & 52.0$\pm$1.1 & 29.0$\pm$0.9  \\
    & &\cellcolor{Gray} {\model w/cos} & \cellcolor{Gray}45.8$\pm$0.2 & \cellcolor{Gray}71.3$\pm$0.3 & \cellcolor{Gray}50.4$\pm$0.3 &\cellcolor{Gray} 50.4$\pm$0.2 & \cellcolor{Gray}77.5$\pm$0.2 &\cellcolor{Gray} 55.9$\pm$0.3 & \cellcolor{Gray}32.0$\pm$0.6 & \cellcolor{Gray}52.8$\pm$1.0 & \cellcolor{Gray}33.7$\pm$0.7  \\ \midrule
\multirow{21}{*}{Split 2} & \multirow{4}{*}{1} & FSRW~\cite{kang2019few} &
    28.4$\pm$0.5&51.7$\pm$0.9&27.3$\pm$0.6&35.7$\pm$0.5&64.8$\pm$0.9&34.6$\pm$0.7&6.3$\pm$0.9&12.3$\pm$1.9&5.5$\pm$0.7 \\
    & & FRCN+ft-full & 30.3$\pm$0.5 & 49.7$\pm$0.5 & 32.3$\pm$0.7 & 38.8$\pm$0.6 & 63.2$\pm$0.7 & 41.6$\pm$0.9 & 5.0$\pm$0.6 & 9.4$\pm$1.2 & 4.5$\pm$0.7 \\
    & & {\model w/fc} & 36.2$\pm$0.8 & 59.6$\pm$0.9 & 38.7$\pm$1.0 & 45.6$\pm$0.9 & 73.8$\pm$0.9 & 49.4$\pm$1.2 & 8.1$\pm$1.2 & 16.9$\pm$2.3 & 6.6$\pm$1.1  \\
    & &\cellcolor{Gray} {\model w/cos} & \cellcolor{Gray}36.7$\pm$0.6 &\cellcolor{Gray} 59.9$\pm$0.8 &\cellcolor{Gray} 39.3$\pm$0.8 &\cellcolor{Gray} 45.9$\pm$0.7 & \cellcolor{Gray}73.8$\pm$0.8 &\cellcolor{Gray} 49.8$\pm$1.1 & \cellcolor{Gray}9.0$\pm$1.2 & \cellcolor{Gray}18.3$\pm$2.4 &\cellcolor{Gray} 7.8$\pm$1.2 \\ \cmidrule{2-12}
    & \multirow{4}{*}{2} & FSRW~\cite{kang2019few} & 
    29.4$\pm$0.3&53.1$\pm$0.6&28.5$\pm$0.4&35.8$\pm$0.4&64.2$\pm$0.6&35.1$\pm$0.5&9.9$\pm$0.7&19.6$\pm$1.3&8.8$\pm$0.6 \\
    & & FRCN+ft-full & 30.7$\pm$0.5 & 49.7$\pm$0.7 & 32.9$\pm$0.6 & 38.4$\pm$0.5 & 61.6$\pm$0.7 & 41.4$\pm$0.7 & 7.7$\pm$0.8 & 13.8$\pm$1.4 & 7.4$\pm$0.8 \\
    & &{\model w/fc} & 38.5$\pm$0.5 & 62.8$\pm$0.6 & 41.2$\pm$0.6 & 46.9$\pm$0.5 & 74.9$\pm$0.5 & 51.2$\pm$0.7 & 13.1$\pm$1.0 & 26.4$\pm$1.9 & 11.3$\pm$1.1  \\
    & & \cellcolor{Gray}{\model w/cos} & \cellcolor{Gray}39.0$\pm$0.4 & \cellcolor{Gray}63.0$\pm$0.5 & \cellcolor{Gray}42.1$\pm$0.6 & \cellcolor{Gray}47.3$\pm$0.4 & \cellcolor{Gray}74.9$\pm$0.4 &\cellcolor{Gray} 51.9$\pm$0.7 &\cellcolor{Gray} 14.1$\pm$0.9 &\cellcolor{Gray} 27.5$\pm$1.6 &\cellcolor{Gray} 12.7$\pm$1.0  \\ \cmidrule{2-12}
    & \multirow{4}{*}{3} & FSRW~\cite{kang2019few} & 29.9$\pm$0.3&53.9$\pm$0.4&29.0$\pm$0.4&35.7$\pm$0.3&63.5$\pm$0.4&35.1$\pm$0.4&12.5$\pm$0.7&25.1$\pm$1.4&10.4$\pm$0.7 \\
    & & FRCN+ft-full & 31.1$\pm$0.3 & 50.1$\pm$0.5 & 33.2$\pm$0.5 & 38.1$\pm$0.4 & 61.0$\pm$0.6 & 41.2$\pm$0.5 & 9.8$\pm$0.9 & 17.4$\pm$1.6 & 9.4$\pm$1.0 \\
    & &{\model w/fc} & 39.4$\pm$0.4 & 64.2$\pm$0.5 & 42.0$\pm$0.5 & 47.5$\pm$0.4 & 75.4$\pm$0.5 & 51.7$\pm$0.6 & 15.2$\pm$0.8 & 30.5$\pm$1.5 & 13.1$\pm$0.8  \\
    & & \cellcolor{Gray}{\model w/cos} & \cellcolor{Gray}40.1$\pm$0.3 &\cellcolor{Gray} 64.5$\pm$0.5 &\cellcolor{Gray} 43.3$\pm$0.4 &\cellcolor{Gray} 48.1$\pm$0.3 & \cellcolor{Gray}75.6$\pm$0.4 & \cellcolor{Gray}52.9$\pm$0.5 & \cellcolor{Gray}16.0$\pm$0.8 & \cellcolor{Gray}30.9$\pm$1.6 & \cellcolor{Gray}14.4$\pm$0.9  \\ \cmidrule{2-12}
    & \multirow{4}{*}{5} & FSRW~\cite{kang2019few} & 30.4$\pm$0.4&54.6$\pm$0.5&29.5$\pm$0.5&35.3$\pm$0.3&62.4$\pm$0.4&34.9$\pm$0.5&15.7$\pm$0.8&31.4$\pm$1.5&13.3$\pm$0.9 \\
    & & FRCN+ft-full & 31.5$\pm$0.3 & 50.8$\pm$0.7 & 33.6$\pm$0.4 & 37.9$\pm$0.4 & 60.4$\pm$0.6 & 40.8$\pm$0.5 & 12.4$\pm$0.9 & 21.9$\pm$1.5 & 12.1$\pm$0.9 \\
    & &{\model w/fc} & 40.0$\pm$0.4 & 65.1$\pm$0.5 & 42.6$\pm$0.5 & 47.5$\pm$0.4 & 75.3$\pm$0.5 & 51.6$\pm$0.5 & 17.5$\pm$0.7 & 34.6$\pm$1.1 & 15.5$\pm$0.9  \\
    & & \cellcolor{Gray}{\model w/cos} & \cellcolor{Gray}40.9$\pm$0.4 & \cellcolor{Gray}65.7$\pm$0.5 & \cellcolor{Gray}44.1$\pm$0.5 & \cellcolor{Gray}48.6$\pm$0.4 &\cellcolor{Gray} 76.2$\pm$0.4 & \cellcolor{Gray}53.3$\pm$0.5 & \cellcolor{Gray}17.8$\pm$0.8 &\cellcolor{Gray} 34.1$\pm$1.4 & \cellcolor{Gray}16.2$\pm$1.0  \\ \cmidrule{2-12}
    & \multirow{3}{*}{10} & FRCN+ft-full & 32.2$\pm$0.3 & 52.3$\pm$0.4 & 34.1$\pm$0.4 & 37.2$\pm$0.3 & 59.8$\pm$0.4 & 39.9$\pm$0.4 & 17.0$\pm$0.8 & 29.8$\pm$1.4 & 16.7$\pm$0.9 \\
    & & {\model w/fc} & 41.3$\pm$0.2 & 67.0$\pm$0.3 & 44.0$\pm$0.3 & 48.3$\pm$0.2 & 76.1$\pm$0.3 & 52.7$\pm$0.4 & 20.2$\pm$0.5 & 39.7$\pm$0.9 & 18.0$\pm$0.7  \\
    & & \cellcolor{Gray}{\model w/cos} & \cellcolor{Gray}42.3$\pm$0.3 &\cellcolor{Gray} 67.6$\pm$0.4 &\cellcolor{Gray} 45.7$\pm$0.3 &\cellcolor{Gray} 49.4$\pm$0.2 & \cellcolor{Gray}76.9$\pm$0.3 & \cellcolor{Gray}54.5$\pm$0.3 &\cellcolor{Gray} 20.8$\pm$0.6 &\cellcolor{Gray} 39.5$\pm$1.1 & \cellcolor{Gray}19.2$\pm$0.6  \\ \midrule
\multirow{21}{*}{Split 3} & \multirow{4}{*}{1} & FSRW~\cite{kang2019few} &
27.5$\pm$0.6&50.0$\pm$1.0&26.8$\pm$0.7&34.5$\pm$0.7&62.5$\pm$1.2&33.5$\pm$0.7&6.7$\pm$1.0&12.5$\pm$1.6&6.4$\pm$1.0 \\
    & & FRCN+ft-full & 30.8$\pm$0.6 & 49.8$\pm$0.8 & 32.9$\pm$0.8 & 39.6$\pm$0.8 & 63.7$\pm$1.0 & 42.5$\pm$0.9 & 4.5$\pm$0.7 & 8.1$\pm$1.3 & 4.2$\pm$0.7 \\
    & & {\model w/fc} & 39.0$\pm$0.6 & 62.3$\pm$0.7 & 42.1$\pm$0.8 & 49.5$\pm$0.8 & 77.8$\pm$0.8 & 54.0$\pm$1.0 & 7.8$\pm$1.1 & 15.7$\pm$2.1 & 6.5$\pm$1.0 \\
    & & \cellcolor{Gray}{\model w/cos} & \cellcolor{Gray}40.1$\pm$0.3 &\cellcolor{Gray} 63.5$\pm$0.6 &\cellcolor{Gray} 43.6$\pm$0.5 &\cellcolor{Gray} 50.2$\pm$0.4 & \cellcolor{Gray}78.7$\pm$0.2 & \cellcolor{Gray}55.1$\pm$0.5 & \cellcolor{Gray}9.6$\pm$1.1 & \cellcolor{Gray}17.9$\pm$2.0 &\cellcolor{Gray} 9.1$\pm$1.2  \\ \cmidrule{2-12}
    & \multirow{4}{*}{2} & FSRW~\cite{kang2019few} & 28.7$\pm$0.4&51.8$\pm$0.7&28.1$\pm$0.5&34.5$\pm$0.4&62.0$\pm$0.7&34.0$\pm$0.5&11.3$\pm$0.7&21.3$\pm$1.0&10.6$\pm$0.8 \\
    & & FRCN+ft-full & 31.3$\pm$0.5 & 50.2$\pm$0.9 & 33.5$\pm$0.6 & 39.1$\pm$0.5 & 62.4$\pm$0.9 & 42.0$\pm$0.7 & 8.0$\pm$0.8 & 13.9$\pm$1.4 & 7.9$\pm$0.9 \\
    & &{\model w/fc} & 41.1$\pm$0.6 & 65.1$\pm$0.7 & 44.3$\pm$0.7 & 50.1$\pm$0.7 & 77.7$\pm$0.7 & 54.8$\pm$0.9 & 14.2$\pm$1.2 & 27.2$\pm$2.0 & 12.6$\pm$1.3  \\
    & & \cellcolor{Gray}{\model w/cos} & \cellcolor{Gray}41.8$\pm$0.4 &\cellcolor{Gray} 65.6$\pm$0.6 &\cellcolor{Gray} 45.3$\pm$0.4 & \cellcolor{Gray}50.7$\pm$0.3 & \cellcolor{Gray}78.4$\pm$0.2 & \cellcolor{Gray}55.6$\pm$0.4 & \cellcolor{Gray}15.1$\pm$1.3 &\cellcolor{Gray} 27.2$\pm$2.1 & \cellcolor{Gray}14.4$\pm$1.5  \\ \cmidrule{2-12}
    & \multirow{4}{*}{3} & FSRW~\cite{kang2019few} & 
    29.2$\pm$0.4&52.7$\pm$0.6&28.5$\pm$0.4&34.2$\pm$0.3&61.3$\pm$0.6&33.6$\pm$0.4&14.2$\pm$0.7&26.8$\pm$1.4&13.1$\pm$0.7 \\
    & & FRCN+ft-full & 32.1$\pm$0.5 & 51.3$\pm$0.8 & 34.3$\pm$0.6 & 39.1$\pm$0.5 & 62.1$\pm$0.7 & 42.1$\pm$0.6 & 11.1$\pm$0.9 & 19.0$\pm$1.5 & 11.2$\pm$1.0 \\
    & &{\model w/fc} & 40.4$\pm$0.5 & 65.4$\pm$0.7 & 43.1$\pm$0.7 & 47.8$\pm$0.5 & 75.6$\pm$0.5 & 52.1$\pm$0.7 & 18.1$\pm$1.0 & 34.7$\pm$1.6 & 16.2$\pm$1.3  \\
    & &\cellcolor{Gray} {\model w/cos} &\cellcolor{Gray} 43.1$\pm$0.4 & \cellcolor{Gray}67.5$\pm$0.5 & \cellcolor{Gray}46.7$\pm$0.5 & \cellcolor{Gray}51.1$\pm$0.3 &\cellcolor{Gray} 78.6$\pm$0.2 &\cellcolor{Gray} 56.3$\pm$0.4 & \cellcolor{Gray}18.9$\pm$1.1 &\cellcolor{Gray} 34.3$\pm$1.7 &\cellcolor{Gray} 18.1$\pm$1.4  \\ \cmidrule{2-12}
    & \multirow{4}{*}{5} & FSRW~\cite{kang2019few} & 
    30.1$\pm$0.3&53.8$\pm$0.5&29.3$\pm$0.4&34.1$\pm$0.3&60.5$\pm$0.4&33.6$\pm$0.4&18.0$\pm$0.7&33.8$\pm$1.4&16.5$\pm$0.8 \\
    & & FRCN+ft-full & 32.4$\pm$0.5 & 51.7$\pm$0.8 & 34.4$\pm$0.6 & 38.5$\pm$0.5 & 61.0$\pm$0.7 & 41.3$\pm$0.6 & 14.0$\pm$0.9 & 23.9$\pm$1.7 & 13.7$\pm$0.9 \\
    & & {\model w/fc} & 41.3$\pm$0.5 & 67.1$\pm$0.6 & 44.0$\pm$0.6 & 48.0$\pm$0.5 & 75.8$\pm$0.5 & 52.2$\pm$0.6 & 21.4$\pm$0.9 & 40.8$\pm$1.3 & 19.4$\pm$1.0  \\
    & & \cellcolor{Gray}{\model w/cos} & \cellcolor{Gray}44.1$\pm$0.3 &\cellcolor{Gray} 69.1$\pm$0.4 & \cellcolor{Gray}47.8$\pm$0.4 & \cellcolor{Gray}51.3$\pm$0.2 & \cellcolor{Gray}78.5$\pm$0.3 & \cellcolor{Gray}56.4$\pm$0.3 & \cellcolor{Gray}22.8$\pm$0.9 & \cellcolor{Gray}40.8$\pm$1.4 & \cellcolor{Gray}22.1$\pm$1.1  \\ \cmidrule{2-12}
    & \multirow{3}{*}{10} & FRCN+ft-full & 33.1$\pm$0.5 & 53.1$\pm$0.7 & 35.2$\pm$0.5 & 38.0$\pm$0.5 & 60.5$\pm$0.7 & 40.7$\pm$0.6 & 18.4$\pm$0.8 & 31.0$\pm$1.2 & 18.7$\pm$1.0 \\
    & & {\model w/fc} & 42.2$\pm$0.4 & 68.3$\pm$0.5 & 44.9$\pm$0.6 & 48.5$\pm$0.4 & 76.2$\pm$0.4 & 52.9$\pm$0.5 & 23.3$\pm$0.8 & 44.6$\pm$1.1 & 21.0$\pm$1.2  \\
    & & \cellcolor{Gray}{\model w/cos} & \cellcolor{Gray}45.0$\pm$0.3 & \cellcolor{Gray}70.3$\pm$0.4 &\cellcolor{Gray} 48.9$\pm$0.4 &\cellcolor{Gray} 51.6$\pm$0.2 &\cellcolor{Gray} 78.6$\pm$0.2 &\cellcolor{Gray} 57.0$\pm$0.3 & \cellcolor{Gray}25.4$\pm$0.7 & \cellcolor{Gray}45.6$\pm$1.1 & \cellcolor{Gray}24.7$\pm$1.1  \\
\bottomrule
\end{tabular}}
\label{tab:voc_bench}
\end{table*}

\begin{table*}[ht]
\centering
\footnotesize
\setlength{\tabcolsep}{0.4em}
\caption{Generalized object detection benchmarks on COCO. For each metric, we report the average and 95\% confidence interval computed over 10 random samples. \vspace{1mm}}
\adjustbox{width=\linewidth}{
\begin{tabular}{c|c|cccccc|ccc|ccc}
\toprule
\multirow{2}{*}{\# shots} &\multirow{2}{*}{Method} &  \multicolumn{6}{c|}{Overall}  &\multicolumn{3}{c|}{Base class} & \multicolumn{3}{c}{Novel class} \\ \cmidrule{3-14}
&  & AP & AP50 & AP75 & APs & APm & APl & bAP & bAP50 & bAP75 & nAP & nAP50 & nAP75 \\ \midrule
\multirow{3}{*}{1} & FRCN+ft-full & 16.2$\pm$0.9 & 25.8$\pm$1.2 & 17.6$\pm$1.0 & 7.2$\pm$0.6 & 17.9$\pm$1.0 & 23.1$\pm$1.1 & 21.0$\pm$1.2 & 33.3$\pm$1.7 & 23.0$\pm$1.4 & 1.7$\pm$0.2 & 3.3$\pm$0.3 & 1.6$\pm$0.2 \\
 & {\model w/fc} & 24.0$\pm$0.5 & 38.9$\pm$0.5 & 25.8$\pm$0.6 & 13.8$\pm$0.4 & 26.6$\pm$0.4 & 32.0$\pm$0.6 & 31.5$\pm$0.5 & 50.7$\pm$0.6 & 33.9$\pm$0.8 & 1.6$\pm$0.4 & 3.4$\pm$0.6 & 1.3$\pm$0.4 \\
 & {\cellcolor{Gray} \model w/cos} &  \cellcolor{Gray}24.4$\pm$0.6 & \cellcolor{Gray}39.8$\pm$0.8 & \cellcolor{Gray}26.1$\pm$0.8 & \cellcolor{Gray}14.7$\pm$0.7 & \cellcolor{Gray}26.8$\pm$0.5 & \cellcolor{Gray}31.4$\pm$0.7 & \cellcolor{Gray}31.9$\pm$0.7 & \cellcolor{Gray}51.8$\pm$0.9 & \cellcolor{Gray}34.3$\pm$0.9 & \cellcolor{Gray}1.9$\pm$0.4 & \cellcolor{Gray}3.8$\pm$0.6 & \cellcolor{Gray}1.7$\pm$0.5 \\ \midrule
\multirow{3}{*}{2} & FRCN+ft-full & 15.8$\pm$0.7 & 25.0$\pm$1.1 & 17.3$\pm$0.7 & 6.6$\pm$0.6 & 17.2$\pm$0.8 & 23.5$\pm$0.7 & 20.0$\pm$0.9 & 31.4$\pm$1.5 & 22.2$\pm$1.0 & 3.1$\pm$0.3 & 6.1$\pm$0.6 & 2.9$\pm$0.3 \\
 & {\model w/fc} & 24.5$\pm$0.4 & 39.3$\pm$0.6 & 26.5$\pm$0.5 & 13.9$\pm$0.3 & 27.1$\pm$0.5 & 32.7$\pm$0.7 & 31.4$\pm$0.5 & 49.8$\pm$0.7 & 34.3$\pm$0.6 & 3.8$\pm$0.5 & 7.8$\pm$0.8 & 3.2$\pm$0.6 \\
 & {\cellcolor{Gray} \model w/cos} & \cellcolor{Gray}24.9$\pm$0.6 & \cellcolor{Gray}40.1$\pm$0.9 & \cellcolor{Gray}27.0$\pm$0.7 & \cellcolor{Gray}14.9$\pm$0.7 & \cellcolor{Gray}27.3$\pm$0.6 & \cellcolor{Gray}32.3$\pm$0.6 & \cellcolor{Gray}31.9$\pm$0.7 & \cellcolor{Gray}50.8$\pm$1.1 & \cellcolor{Gray}34.8$\pm$0.8 & \cellcolor{Gray}3.9$\pm$0.4 & \cellcolor{Gray}7.8$\pm$0.7 & \cellcolor{Gray}3.6$\pm$0.6 \\ \midrule
\multirow{3}{*}{3} & FRCN+ft-full & 15.0$\pm$0.7 & 23.9$\pm$1.2 & 16.4$\pm$0.7 & 6.0$\pm$0.6 & 16.1$\pm$0.9 & 22.6$\pm$0.9 & 18.8$\pm$0.9 & 29.5$\pm$1.5 & 20.7$\pm$0.9 & 3.7$\pm$0.4 & 7.1$\pm$0.8 & 3.5$\pm$0.4 \\
 & {\model w/fc} & 24.9$\pm$0.5 & 39.7$\pm$0.7 & 27.1$\pm$0.6 & 14.1$\pm$0.4 & 27.5$\pm$0.6 & 33.4$\pm$0.8 & 31.5$\pm$0.6 & 49.6$\pm$0.7 & 34.6$\pm$0.7 & 5.0$\pm$0.5 & 9.9$\pm$1.0 & 4.6$\pm$0.6 \\
 & {\cellcolor{Gray} \model w/cos} & \cellcolor{Gray}25.3$\pm$0.6 & \cellcolor{Gray}40.4$\pm$1.0 & \cellcolor{Gray}27.6$\pm$0.7 & \cellcolor{Gray}14.8$\pm$0.7 & \cellcolor{Gray}27.7$\pm$0.6 & \cellcolor{Gray}33.1$\pm$0.7 & \cellcolor{Gray}32.0$\pm$0.7 & \cellcolor{Gray}50.5$\pm$1.0 & \cellcolor{Gray}35.1$\pm$0.7 & \cellcolor{Gray}5.1$\pm$0.6 & \cellcolor{Gray}9.9$\pm$0.9 & \cellcolor{Gray}4.8$\pm$0.6 \\ \midrule
\multirow{3}{*}{5} & FRCN+ft-full & 14.4$\pm$0.8 & 23.0$\pm$1.3 & 15.6$\pm$0.8 & 5.6$\pm$0.4 & 15.2$\pm$1.0 & 21.9$\pm$1.1 & 17.6$\pm$0.9 & 27.8$\pm$1.5 & 19.3$\pm$1.0 & 4.6$\pm$0.5 & 8.7$\pm$1.0 & 4.4$\pm$0.6 \\
 & {\model w/fc} & 25.6$\pm$0.5 & 40.7$\pm$0.8 & 28.0$\pm$0.5 & 14.3$\pm$0.4 & 28.2$\pm$0.6 & 34.4$\pm$0.6 & 31.8$\pm$0.5 & 49.8$\pm$0.7 & 35.2$\pm$0.5 & 6.9$\pm$0.7 & 13.4$\pm$1.2 & 6.3$\pm$0.8 \\
 & {\cellcolor{Gray} \model w/cos} & \cellcolor{Gray}25.9$\pm$0.6 & \cellcolor{Gray}41.2$\pm$0.9 & \cellcolor{Gray}28.4$\pm$0.6 & \cellcolor{Gray}15.0$\pm$0.6 & \cellcolor{Gray}28.3$\pm$0.5 & \cellcolor{Gray}34.1$\pm$0.6 & \cellcolor{Gray}32.3$\pm$0.6 & \cellcolor{Gray}50.5$\pm$0.9 & \cellcolor{Gray}35.6$\pm$0.6 & \cellcolor{Gray}7.0$\pm$0.7 & \cellcolor{Gray}13.3$\pm$1.2 & \cellcolor{Gray}6.5$\pm$0.7 \\ \midrule
\multirow{3}{*}{10} & FRCN+ft-full & 13.4$\pm$1.0 & 21.8$\pm$1.7 & 14.5$\pm$0.9 & 5.1$\pm$0.4 & 14.3$\pm$1.2 & 20.1$\pm$1.5 & 16.1$\pm$1.0 & 25.7$\pm$1.8 & 17.5$\pm$1.0 & 5.5$\pm$0.9 & 10.0$\pm$1.6 & 5.5$\pm$0.9 \\
 & {\model w/fc} & 26.2$\pm$0.5 & 41.8$\pm$0.7 & 28.6$\pm$0.5 & 14.5$\pm$0.3 & 29.0$\pm$0.5 & 35.2$\pm$0.6 & 32.0$\pm$0.5 & 49.9$\pm$0.7 & 35.3$\pm$0.6 & 9.1$\pm$0.5 & 17.3$\pm$1.0 & 8.5$\pm$0.5 \\
 & {\cellcolor{Gray} \model w/cos} & \cellcolor{Gray}26.6$\pm$0.5 & \cellcolor{Gray}42.2$\pm$0.8 & \cellcolor{Gray}29.0$\pm$0.6 & \cellcolor{Gray}15.0$\pm$0.5 & \cellcolor{Gray}29.1$\pm$0.4 & \cellcolor{Gray}35.2$\pm$0.5 & \cellcolor{Gray}32.4$\pm$0.6 & \cellcolor{Gray}50.6$\pm$0.9 & \cellcolor{Gray}35.7$\pm$0.7 & \cellcolor{Gray}9.1$\pm$0.5 & \cellcolor{Gray}17.1$\pm$1.1 & \cellcolor{Gray}8.8$\pm$0.5 \\ \midrule
\multirow{3}{*}{30} & FRCN+ft-full & 13.5$\pm$1.0 & 21.8$\pm$1.9 & 14.5$\pm$1.0 & 5.1$\pm$0.3 & 14.6$\pm$1.2 & 19.9$\pm$2.0 & 15.6$\pm$1.0 & 24.8$\pm$1.8 & 16.9$\pm$1.0 & 7.4$\pm$1.1 & 13.1$\pm$2.1 & 7.4$\pm$1.0 \\
 & {\model w/fc} & 28.4$\pm$0.3 & 44.4$\pm$0.6 & 31.2$\pm$0.3 & 15.7$\pm$0.3 & 31.2$\pm$0.3 & 38.6$\pm$0.4 & 33.8$\pm$0.3 & 51.8$\pm$0.6 & 37.6$\pm$0.4 & 12.0$\pm$0.4 & 22.2$\pm$0.6 & 11.8$\pm$0.4 \\
 & {\cellcolor{Gray} \model w/cos} & \cellcolor{Gray}28.7$\pm$0.4 & \cellcolor{Gray}44.7$\pm$0.7 & \cellcolor{Gray}31.5$\pm$0.4 & \cellcolor{Gray}16.1$\pm$0.4 & \cellcolor{Gray}31.2$\pm$0.3 & \cellcolor{Gray}38.4$\pm$0.4 & \cellcolor{Gray}34.2$\pm$0.4 & \cellcolor{Gray}52.3$\pm$0.7 & \cellcolor{Gray}38.0$\pm$0.4 & \cellcolor{Gray}12.1$\pm$0.4 & \cellcolor{Gray}22.0$\pm$0.7 & \cellcolor{Gray}12.0$\pm$0.5 \\
\bottomrule
\end{tabular}}
\vspace{-1mm}
\label{tab:coco_bench}
\end{table*}

\section{Performance over Multiple Runs}
In our revised benchmark, we adopt $n$ repeated runs with different randomly sampled training shots to increase the reliability of the benchmark. In our experiments, we adopt $n=30$ for PASCAL VOC and $n=10$ for COCO. 

In this section, we provide plots of cumulative means with 95\% confidence intervals of the repeated runs to show
that the selected value of $n$ is sufficient to provide statistically stable results. 

\begin{figure*}[ht]
	\begin{center}
		\centerline{\includegraphics[width=\columnwidth*2]{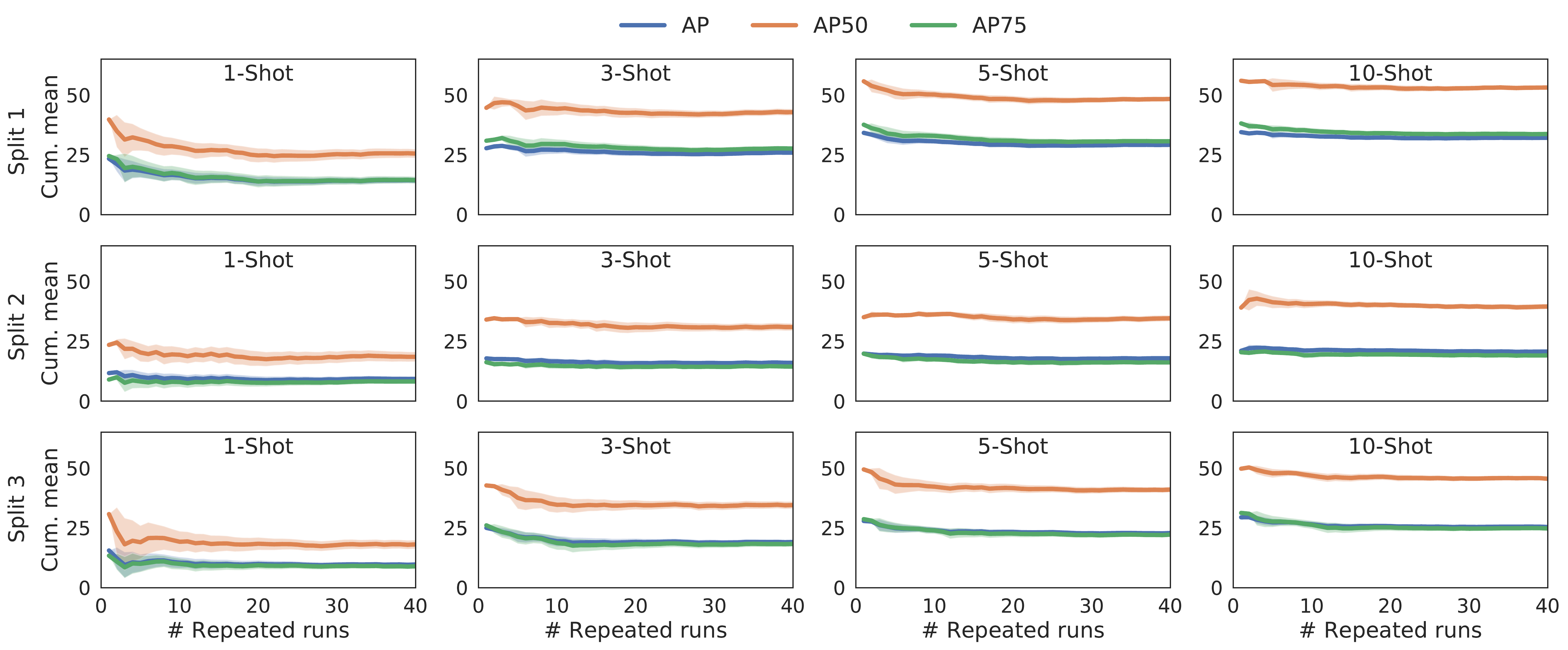}}
		\vspace{-5mm}
		\caption{Cumulative means with 95\% confidence intervals across 40 repeated runs, computed on the novel classes of all three splits of PASCAL VOC. The means and variances become stable after around 30 runs.}
		\label{fig:avg-ap-sup}
	\end{center}
	\vspace{-10mm}
\end{figure*}

\begin{figure*}[ht]
	\begin{center}
		\centerline{\includegraphics[width=.8\linewidth]{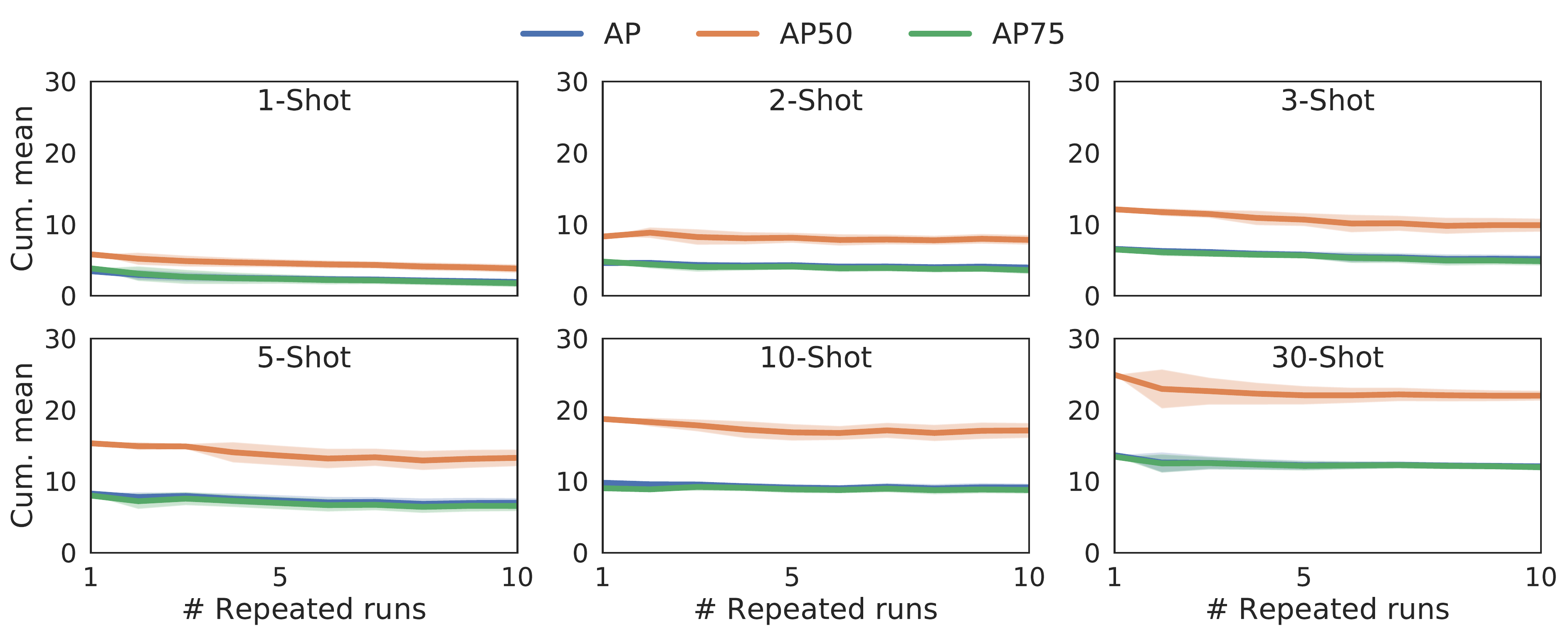}}
		\vspace{-5mm}
		\caption{Cumulative means with 95\% confidence intervals across 10 repeated runs, computed on the novel classes of COCO.}
		\label{fig:coco-avg-ap-sup}
	\end{center}
	\vspace{-10mm}
\end{figure*}

We plot the model performance measured by AP, AP50 and AP75 of up to 40 random groups of training shots across all three splits in Figure~\ref{fig:avg-ap-sup}. For COCO, we plot  up  to  10  random  groups of training shots in Figure~\ref{fig:coco-avg-ap-sup}.

As we can observe from both Figure~\ref{fig:avg-ap-sup} and Figure~\ref{fig:coco-avg-ap-sup}, after around 30 runs on PASCAL VOC and 8 runs on COCO, the means and variances stabilize and our selected values of $n$ are sufficient to obtain stable estimates of the model performances and reliable comparisons across different methods. 

We also observe that the average value across multiple runs is consistently lower than that on the first run, especially in the one-shot case. For example, the average AP50 across 40 runs is around 15 points lower than the AP50 on the first run in the 1-shot case on split 1 on PASCAL VOC. This indicates that the accuracies on the first run, adopted by the previous work~\cite{kang2019few, yan2019meta, wang2019meta}, often overestimate the actual performance and thus lead to unreliable comparison between different approaches.

\end{document}